%% file: ICL2025.tex
\documentclass[a4paper, conference]{IEEEtran}

\usepackage{cite}
\usepackage{amsmath,amssymb,amsfonts}
\usepackage{algorithmic}
\usepackage{graphicx}
\usepackage{textcomp}
\usepackage{xcolor}
\usepackage{paralist}
\usepackage{eurosym}
\usepackage{flushend}
\usepackage{url}
\usepackage[utf8]{inputenc}
\usepackage[colorlinks=true]{hyperref}
\usepackage{soul}
\usepackage{caption}
\usepackage[list=true]{subcaption}
\usepackage{booktabs}
\usepackage{tabu}
\usepackage{wrapfig}
\usepackage{arydshln}
\usepackage[normalem]{ulem}
\usepackage{multirow}
\usepackage{siunitx}
\usepackage{adjustbox}
\usepackage{array}
\usepackage{diagbox}

\def\rot{\rotatebox}

\begin{document}

\title{Evaluating ML Robustness in GNSS Interference Classification, Characterization \& Localization}

\input{00authors.tex}

\IEEEoverridecommandlockouts
\IEEEpubid{\makebox[\columnwidth]{
979-8-3315-1113-5/25/\$31.00~\copyright2025
IEEE \hfill} \hspace{\columnsep}\makebox[\columnwidth]{ }}

\maketitle

\input{00abstract.tex}
\IEEEpeerreviewmaketitle

\input{01introduction.tex}
\input{02related_work.tex}
\input{03experiments.tex}
\input{04method.tex}
\input{05evaluation.tex}
\input{06conclusion.tex}

\section*{Acknowledgments}
\small This work has been carried out within the DARCII project, funding code 50NA2401, sponsored by the German Federal Ministry for Economic Affairs and Climate Action (BMWK) and supported by the German Aerospace Center (DLR), the Bundesnetzagentur (BNetzA), and the Federal Agency for Cartography and Geodesy (BKG).

\bibliography{ICL2025}
\bibliographystyle{IEEEtran}

\end{document}

%% file: 00authors.tex
\author{\IEEEauthorblockN{Lucas Heublein\IEEEauthorrefmark{1},
    Tobias Feigl\IEEEauthorrefmark{1},
    Thorsten Nowak\IEEEauthorrefmark{2},
    Alexander Rügamer\IEEEauthorrefmark{1},
    Christopher Mutschler\IEEEauthorrefmark{1},
    \underline{Felix Ott}\IEEEauthorrefmark{1}}
  \IEEEauthorblockA{\IEEEauthorrefmark{1}Fraunhofer Institute for Integrated Circuits IIS, 90411 Nürnberg, Germany}
  \IEEEauthorblockA{\IEEEauthorrefmark{2}Diehl Defence GmbH \& Co. KG, 90552 Röthenbach an der Pegnitz, Germany}
  \IEEEauthorblockA{\{lucas.heublein, tobias.feigl, alexander.ruegamer, christopher.mutschler, felix.ott\}@iis.fraunhofer.de}
  \IEEEauthorblockA{thorsten.nowak@diehl-defence.com}
  }

%% file: 00abstract.tex
\begin{abstract}
Jamming devices disrupt signals from the global navigation satellite system (GNSS) and pose a significant threat, as they compromise the robustness of accurate positioning. The detection of anomalies within frequency snapshots is crucial to counteract these interferences effectively. A critical preliminary countermeasure involves the reliable classification of interferences and the characterization and localization of jamming devices. This paper introduces an extensive dataset comprising snapshots obtained from a low-frequency antenna that capture various generated interferences within a large-scale environment, including controlled multipath effects. Our objective is to assess the resilience of machine learning (ML) models against environmental changes, such as multipath effects, variations in interference attributes, such as interference class, bandwidth, and signal power, the accuracy of jamming device localization, and the constraints imposed by snapshot input lengths. Furthermore, we evaluate the performance of a diverse set of 129 distinct vision encoder models across all tasks. By analyzing the aleatoric and epistemic uncertainties, we demonstrate the adaptability of our model in generalizing across diverse facets, thus establishing its suitability for real-world applications. Dataset: \href{https://gitlab.cc-asp.fraunhofer.de/darcy_gnss/controlled_low_frequency}{https://gitlab.cc-asp.fraunhofer.de/darcy\_gnss/controlled\_low\_frequency}
\end{abstract}

%% file: 01introduction.tex
\section{Introduction}
\label{label_introduction}

Anomaly detection involves the identification of deviations within data patterns from the expected standard. Within the domain of GNSS-based applications~\cite{raichur_ion_gnss,jdidi_brieger,ott_heublein,raichur_heublein,gaikwad_heublein,manjunath_heublein}, the detection of interference assumes a pivotal role~\cite{heublein_raichur_ion}. The accuracy of GNSS receivers' localization is significantly compromised by interference signals emanating from jammers~\cite{crespillo_ruiz,yuan_shen}. This issue has significantly intensified in recent years (see the latest newspaper \cite{ainonline}) due to the increased prevalence of cost-effective and easily accessible jamming devices \cite{merwe_franco}. Consequently, the imperative task include the detection, classification, characterization, localization, and mitigation of potential interference signals \cite{brieger_ion_gnss}. Given the unpredictable emergence of novel jammer types, the objective is to develop resilient ML models. The data may exhibit substantial variance in jammer devices, interference characteristics, antenna changes, and environmental alterations. Therefore, it is paramount to assess the robustness of ML models in terms of their generalization capabilities against these variables. Ensuring a dependable deployment of ML models in real-world scenarios is of utmost importance.

\begin{figure}[!b]
    \centering
    \includegraphics[width=1.0\linewidth]{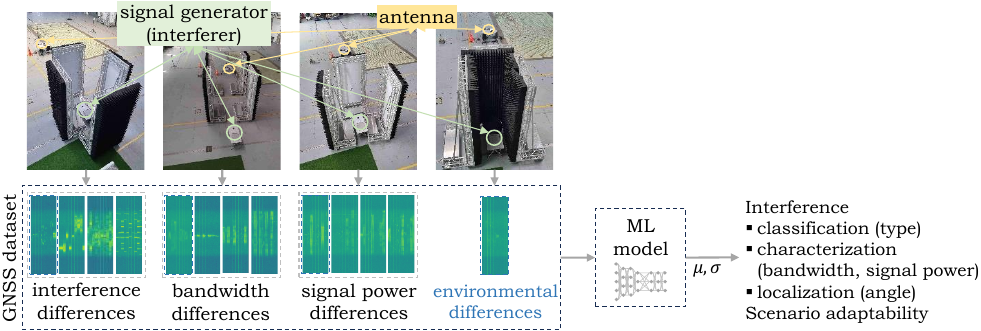}
    \vspace{-0.4cm}
    \caption{Pipeline of evaluating ML robustness for GNSS interference characterization and localization and the adaptability to different multipath scenarios.}
    \label{figure_introduction}
\end{figure}

Assessing the robustness of a model stands as a critical aspect, typically necessitating its application across multiple independent datasets. Given the constraints associated with collecting extensive datasets, attention is often directed towards analyzing the distributions wherein the algorithm exhibits suboptimal performance \cite{subbaswamy_adam,piratla}. Notably, shifts in datasets between training and testing sets commonly lead to a decline in model efficacy, prompting efforts to alleviate such distributional discrepancies~\cite{thamas_oberst,linden_forsberg,wang_gong,heublein_kocher}. One viable approach involves evaluating either the latent space or the predictive uncertainty~\cite{wagh_wei}. However, model robustness is often assessed on synthetic datasets, leaving uncertainties regarding the model's resilience in the face of distributional shifts observed in authentic data scenarios \cite{taori_dave}. Numerous real-world datasets exhibit such distributional shifts, evident in domains like handwriting recognition~\cite{ott_acmmm} and EEG signal classification~\cite{wagh_wei}, where the origins of data shifts are multifaceted. 

The recording of datasets within controlled settings~\cite{ott_fusing} facilitates an exploration of the limitations of ML models. In the context of GNSS, particular emphasis is placed on the detection of non-line-of-sight~\cite{crespillo_ruiz} and multipath~\cite{yuan_shen} signals within specific environments \cite{koiloth_achanta}. However, datasets suitable for certain applications are often sparse, synthetic, or not publicly available. Hence, our approach involves the evaluation of our models using data acquired in controlled environments, with and without multipath effects.

\textbf{Contributions.} The primary objective of this work is to assess the robustness of ML models when applied to GNSS data. Figure~\ref{figure_introduction} illustrates the pipeline employed. Our contributions are outlined as follows: (1) We introduce a GNSS dataset collected within a large-scale industrial setting. To emulate multipath effects, we situate a GNSS antenna and a jamming device (i.e., a signal generator) in a controlled environment, incorporating absorber walls between the antenna and the jammer. (2) Various classification and regression tasks are defined to evaluate the efficacy of ML models. These tasks encompass the classification of interference type, the characterization of interference parameters such as bandwidth (BW) and signal power, and the localization of jammer source. (3) We evaluate the robustness of model predictions amidst environmental variations, notably the generalization from minimal absorption to near-complete absorption of jamming signals. (4) The reliability of model predictions is assessed through the computation of both aleatoric and epistemic uncertainty.

\textbf{Outlook.} The remainder of this paper is organized as follows. Section~\ref{label_related_work} provides an overview of the existing literature for the classification of GNSS interference and the evaluation of the robustness of ML. In Section~\ref{label_experiments}, we introduce our interference snapshot dataset, and Section~\ref{label_method} shows our ML methodology. Section~\ref{label_evaluation} summarizes the evaluation results, followed by the concluding remarks in Section~\ref{label_conclusion}.

%% file: 02related_work.tex
\section{Related Work}
\label{label_related_work}

\textbf{GNSS Interference Classification.} Classical approaches utilize specific features, including automatic gain control (AGC)~\cite{yang_kang}, received signal strength (RSS) values combined with time difference of arrival techniques~\cite{dempster_cetin}, particle filters~\cite{biswas_cetin}, and conformal seven-element antenna arrays~\cite{marcos_caizzone}, as well as methods designed for low-earth orbit applications~\cite{murrian_narula}. In contrast, several recent ML-based approaches, such as convolutional neural networks (CNNs), support vector machines (SVMs), and random forests, have been proposed~\cite{swinney_woods,ferre_fuente,li_huang_lang,xu_ying_li,mehr_dovis,ding_pham}. Ott et al.~\cite{ott_heublein,ott_heublein_jispin} proposed an uncertainty-based Quadruplet loss aiming a more continuous representation between positive and negative interference pairs. They utilize a dataset resembling a snapshot-based real-world dataset, featuring similar interference classes, however, with varying sampling rates. Raichur et al.~\cite{raichur_heublein} used the same dataset to adapt to novel interference classes through continual learning. Jdidi et al.~\cite{jdidi_brieger} proposed an unsupervised method of adapting to diverse environment-specific factors, i.e., multipath effects, dynamics, and variations in signal strength. Raichur et al.~\cite{raichur_ion_gnss} introduced a crowdsourcing method utilizing smartphone-based features to localize the source of any detected interference. Brieger et al.~\cite{brieger_ion_gnss} integrated both the spatial and temporal relationships between samples using a joint loss function and a late fusion technique. Their dataset was acquired in a controlled indoor environment. As ResNet18~\cite{he_zhang} proved to be robust for interference classification, we also employ ResNet18 for feature extraction. 

\textbf{ML Model Robustness.} Koiloth et al.~\cite{koiloth_achanta} benchmarked 14 ML methods to investigate the analysis of multipath data induced by sea waves, however, solely encompassed the consideration of elevation angle, signal strength, and pseudorange residuals. Yuan et al.~\cite{yuan_shen} focused on alleviating estimated multipath biases. Crespillo et al.~\cite{crespillo_ruiz} proposed a logistic regression approach to mitigate biased estimation, thereby enhancing ML generalization. In assessing model robustness, rather than merely detecting outliers within datasets \cite{linden_forsberg,piratla}, it is imperative to analyze the distributions among evaluation data~\cite{subbaswamy_adam}. While Taori et al.~\cite{taori_dave} evaluated the distributional shift from synthetic to real data, our evaluation focuses on assessing model robustness across different scenarios within a controlled environment. Similar to Wagh et al.~\cite{wagh_wei}, we analyze the predictive uncertainty between dataset shifts.

%% file: 03experiments.tex
\section{Experiments}
\label{label_experiments}

\subsection{Dataset}
\label{label_experiments_dataset}

\newcommand\x{0.185}
\begin{figure}[!t]
    \centering
	\begin{minipage}[t]{\x\linewidth}
        \centering
    	\includegraphics[width=1.0\linewidth]{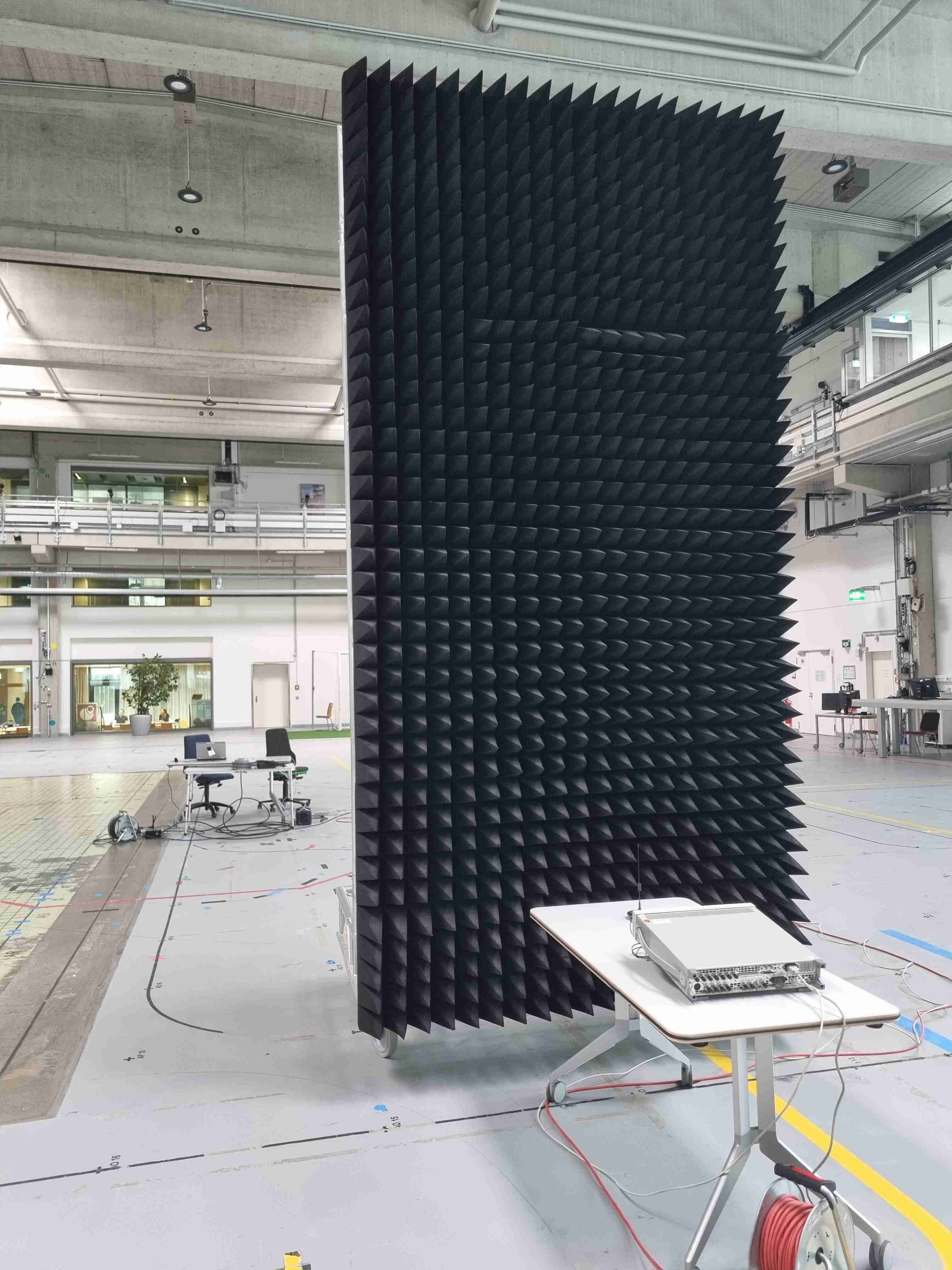}
        \vspace{-0.5cm}
    	\subcaption{Scen. 2.}
    	\label{figure_multi_path2}
    \end{minipage}
    \hfill
	\begin{minipage}[t]{\x\linewidth}
        \centering
    	\includegraphics[width=1.0\linewidth]{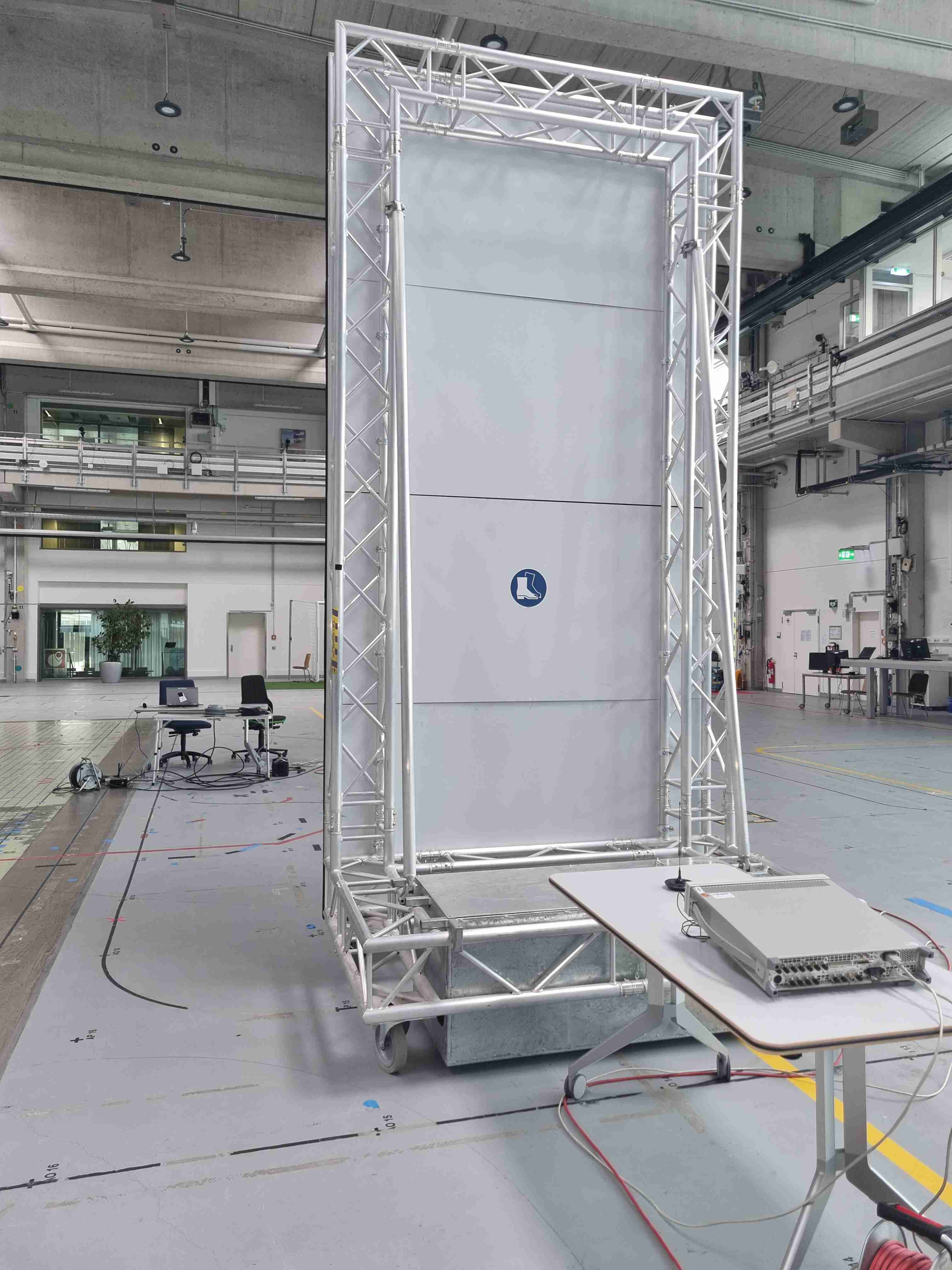}
        \vspace{-0.5cm}
    	\subcaption{Scen. 3.}
    	\label{figure_multi_path3}
    \end{minipage}
    \hfill
	\begin{minipage}[t]{\x\linewidth}
        \centering
    	\includegraphics[width=1.0\linewidth]{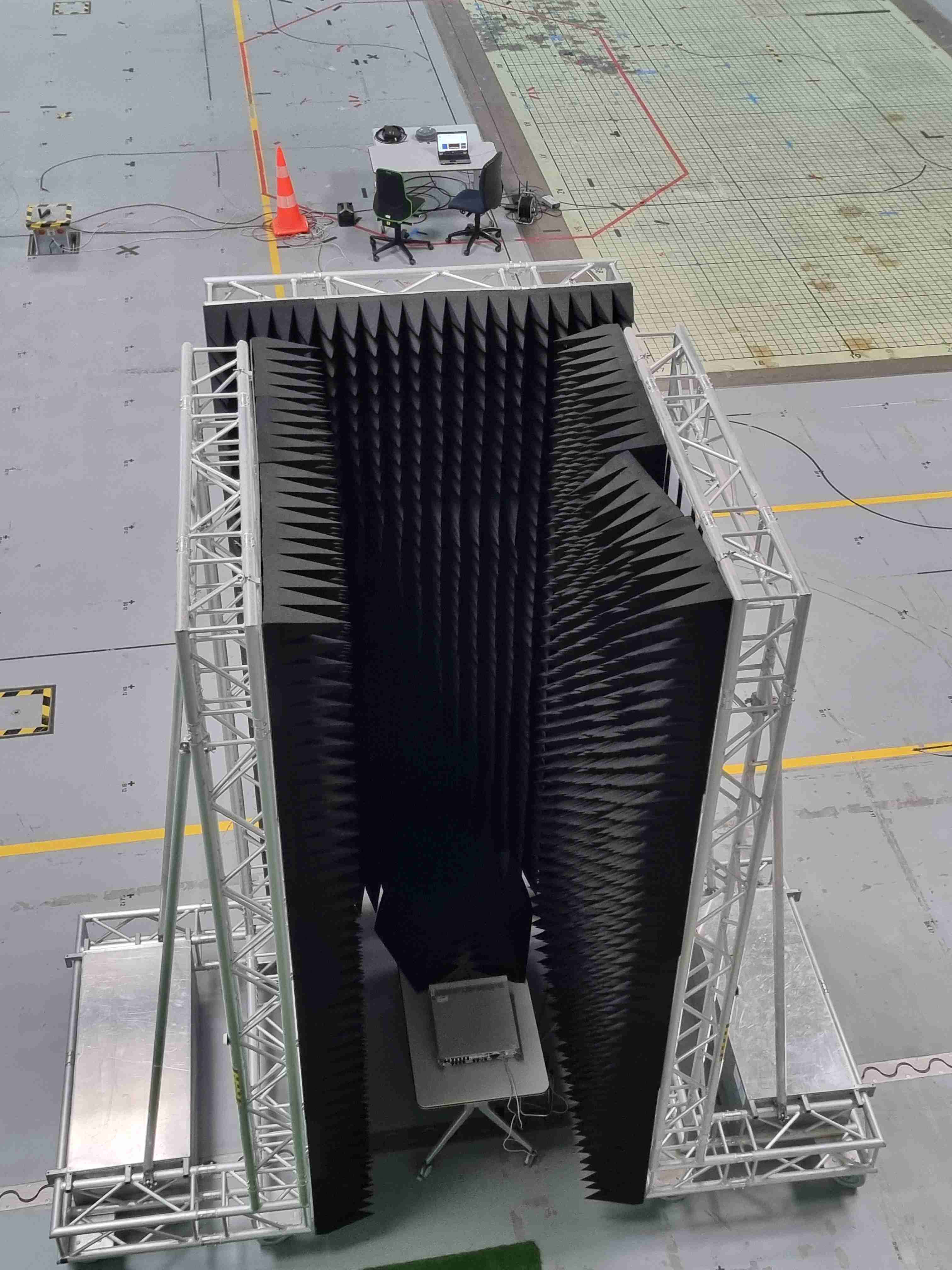}
        \vspace{-0.5cm}
    	\subcaption{Scen. 4.}
    	\label{figure_multi_path4}
    \end{minipage}
    \hfill
	\begin{minipage}[t]{\x\linewidth}
        \centering
    	\includegraphics[width=1.0\linewidth]{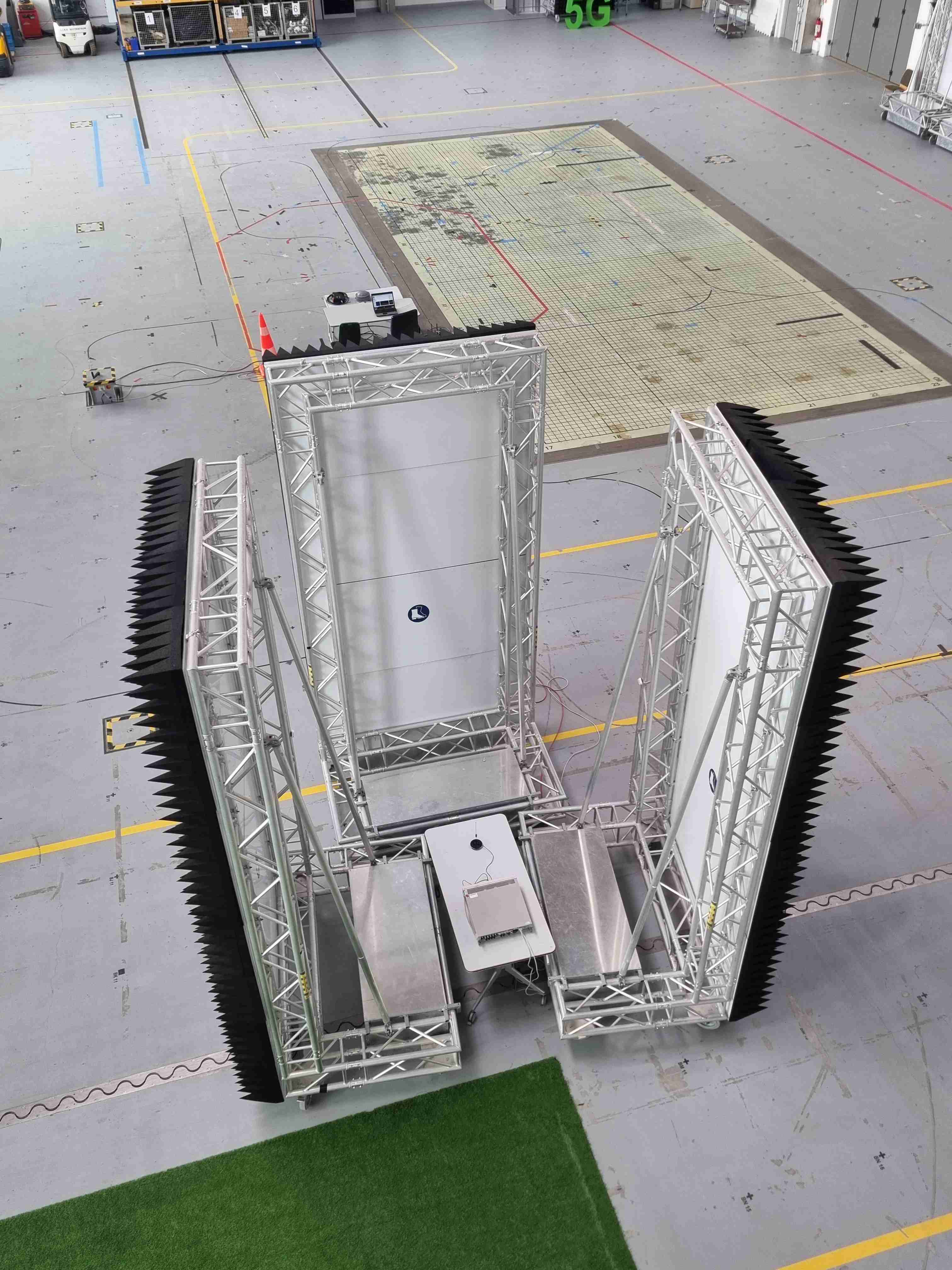}
        \vspace{-0.5cm}
    	\subcaption{Scen. 5.}
    	\label{figure_multi_path5}
    \end{minipage}
    \hfill
	\begin{minipage}[t]{\x\linewidth}
        \centering
    	\includegraphics[width=1.0\linewidth]{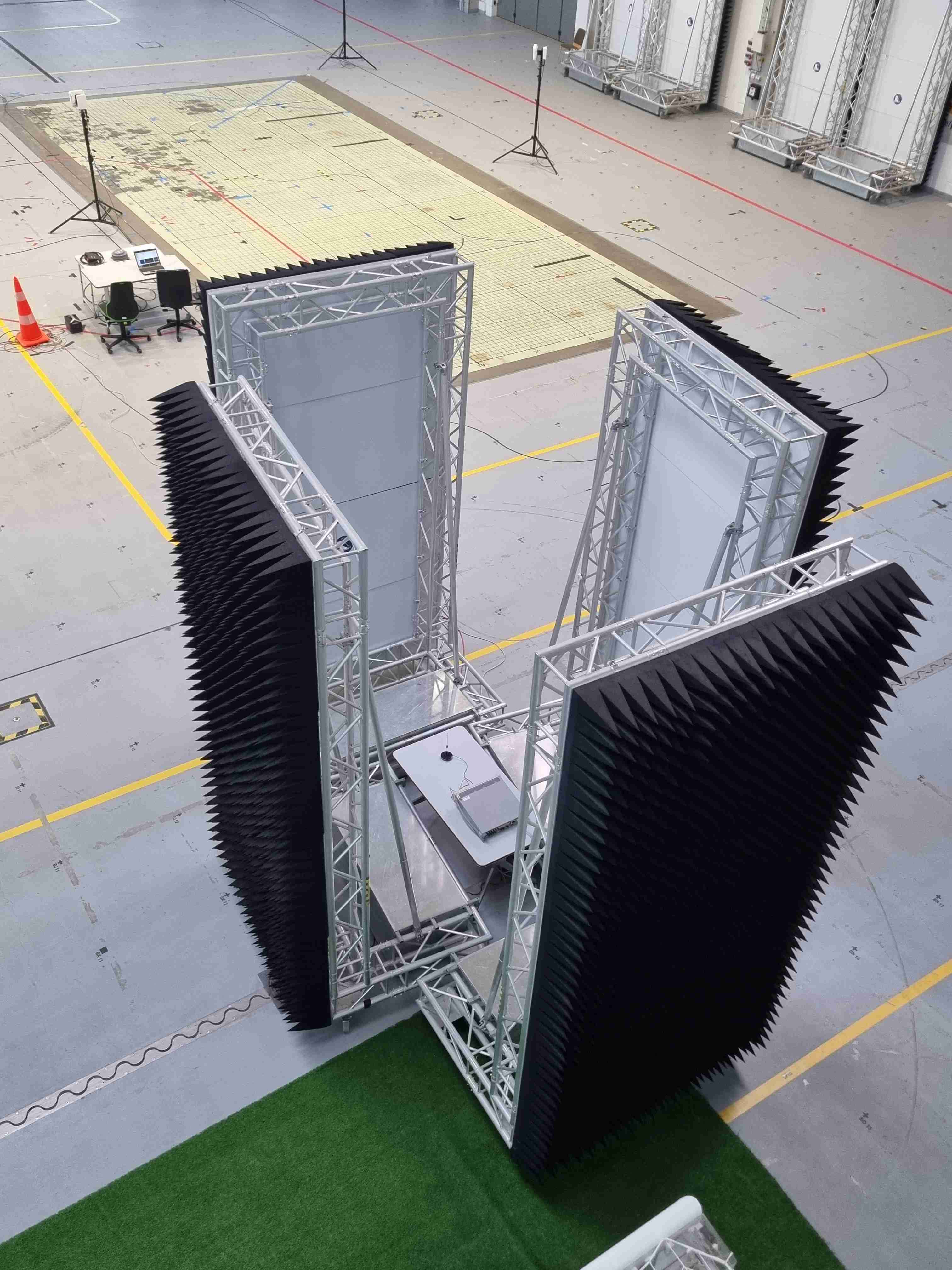}
        \vspace{-0.5cm}
    	\subcaption{Scen. 6.}
    	\label{figure_multi_path6}
    \end{minipage}
	\begin{minipage}[t]{\x\linewidth}
        \centering
    	\includegraphics[width=1.0\linewidth]{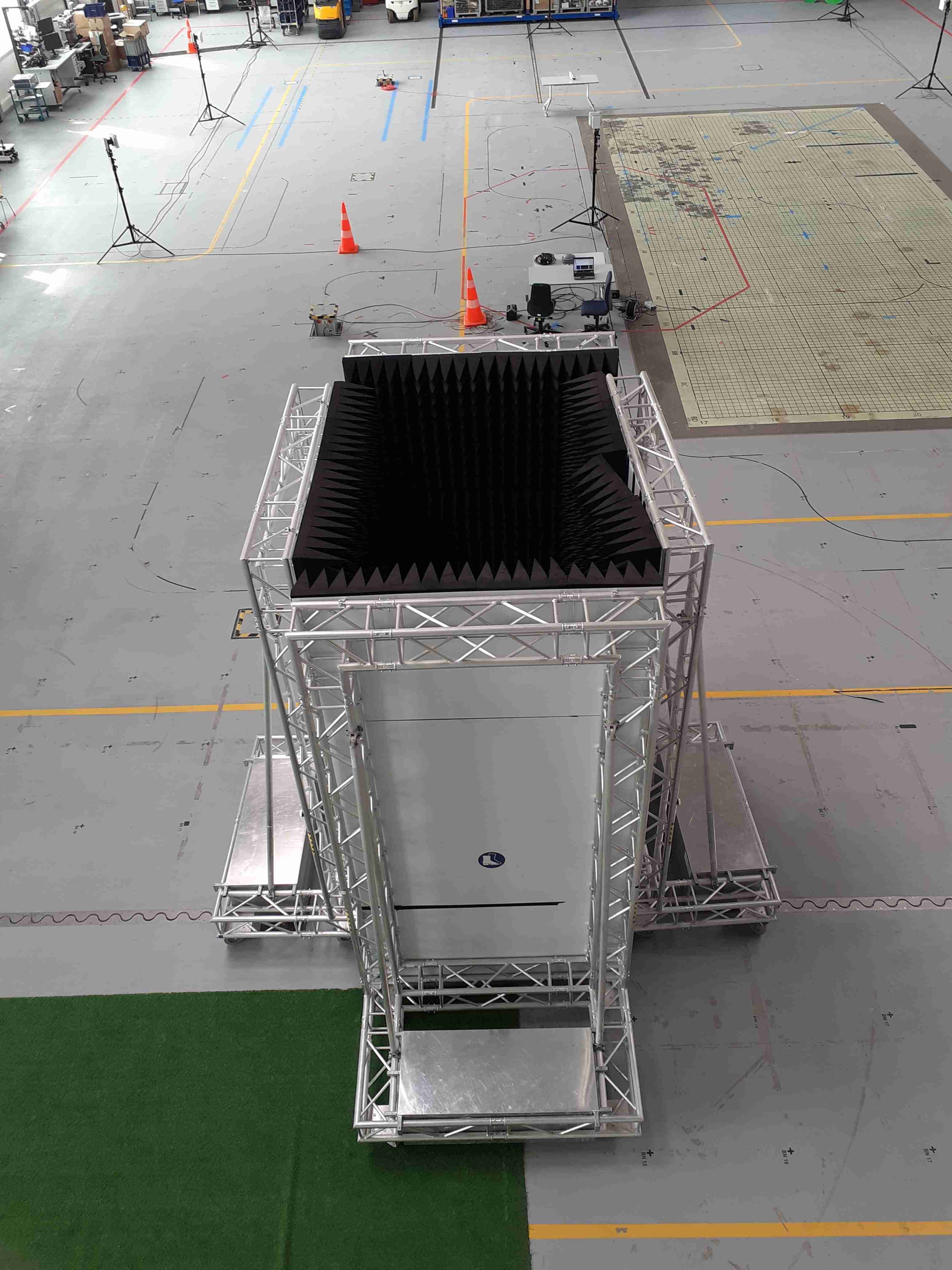}
        \vspace{-0.5cm}
    	\subcaption{Scen. 7.}
    	\label{figure_multi_path7}
    \end{minipage}
    \hfill
	\begin{minipage}[t]{\x\linewidth}
        \centering
    	\includegraphics[width=1.0\linewidth]{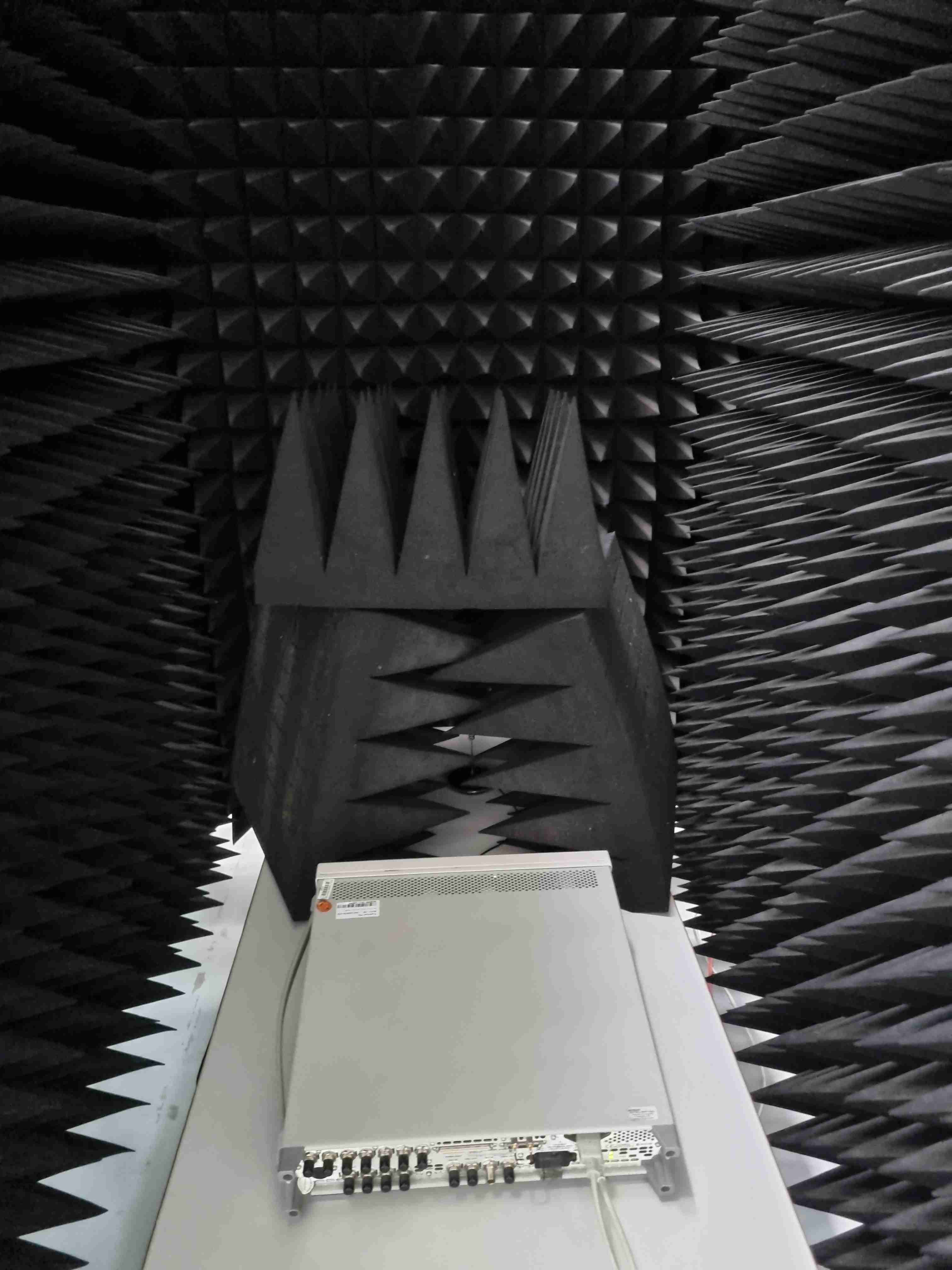}
        \vspace{-0.5cm}
    	\subcaption{Scen. 8.}
    	\label{figure_multi_path8}
    \end{minipage}
    \hfill
	\begin{minipage}[t]{\x\linewidth}
        \centering
    	\includegraphics[width=1.0\linewidth]{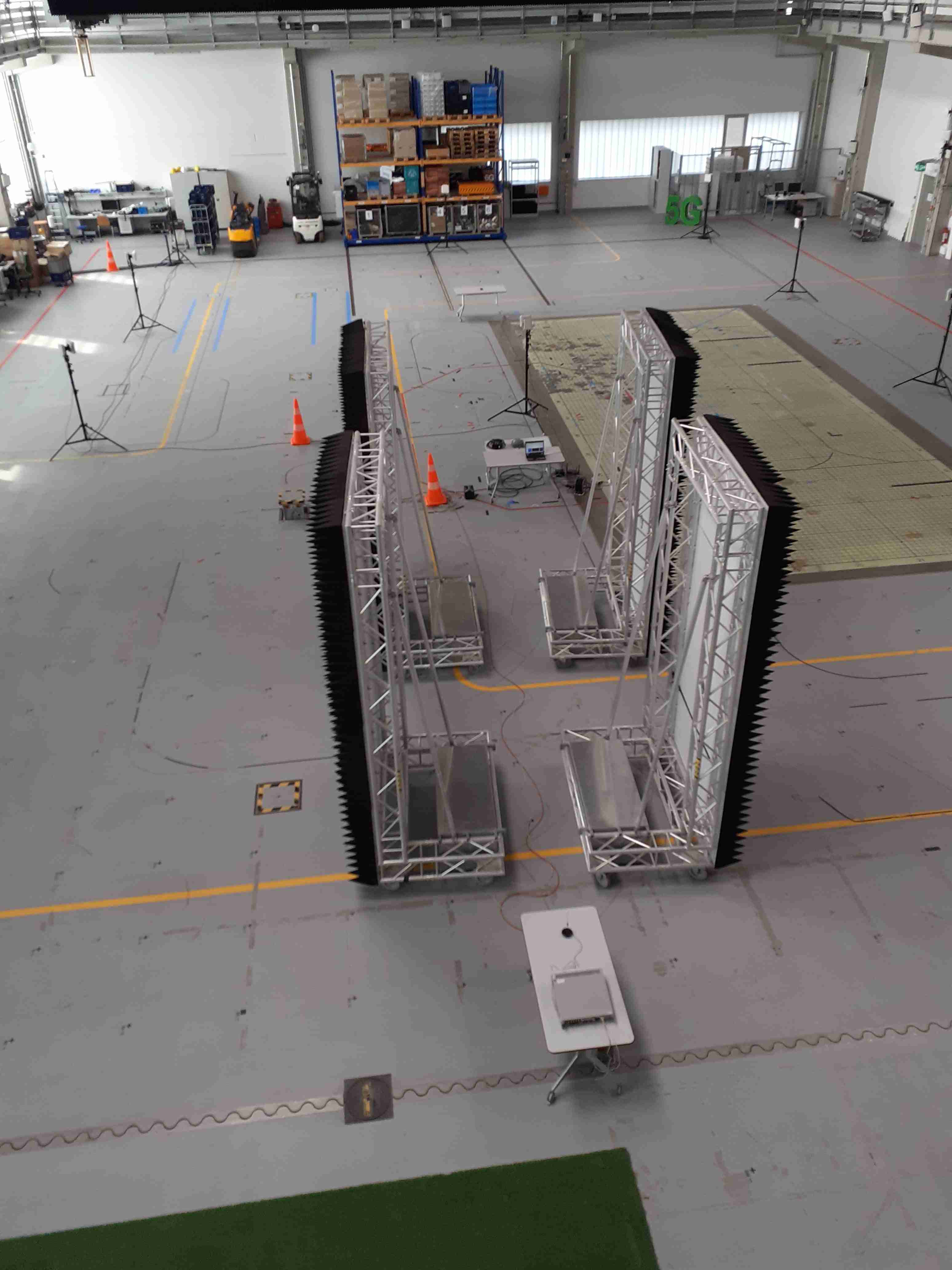}
        \vspace{-0.5cm}
    	\subcaption{Scen. 9.}
    	\label{figure_multi_path9}
    \end{minipage}
    \hfill
	\begin{minipage}[t]{\x\linewidth}
        \centering
    	\includegraphics[width=1.0\linewidth]{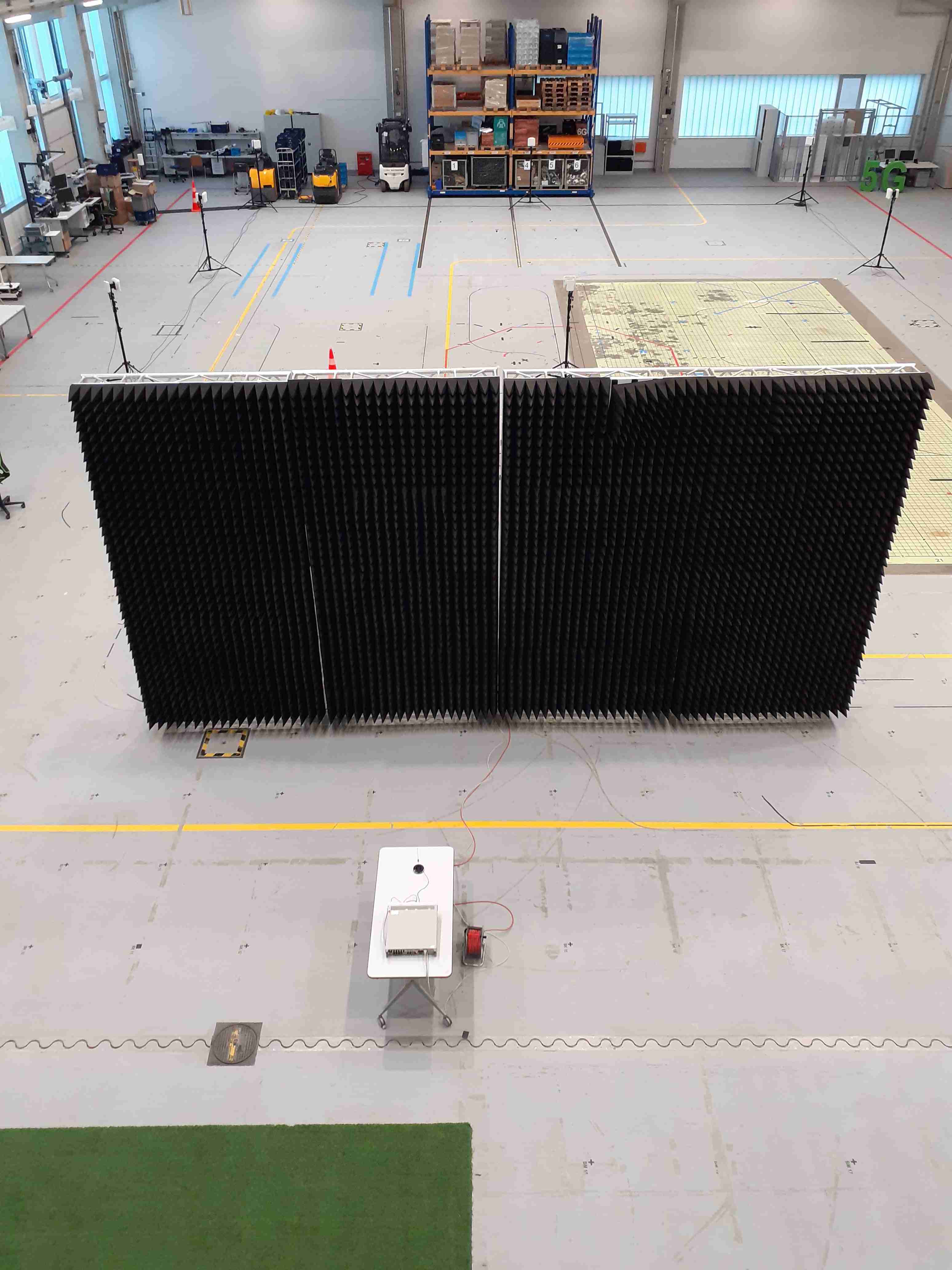}
        \vspace{-0.5cm}
    	\subcaption{Scen. 10.}
    	\label{figure_multi_path10}
    \end{minipage}
    \hfill
	\begin{minipage}[t]{\x\linewidth}
        \centering
    	\includegraphics[width=1.0\linewidth]{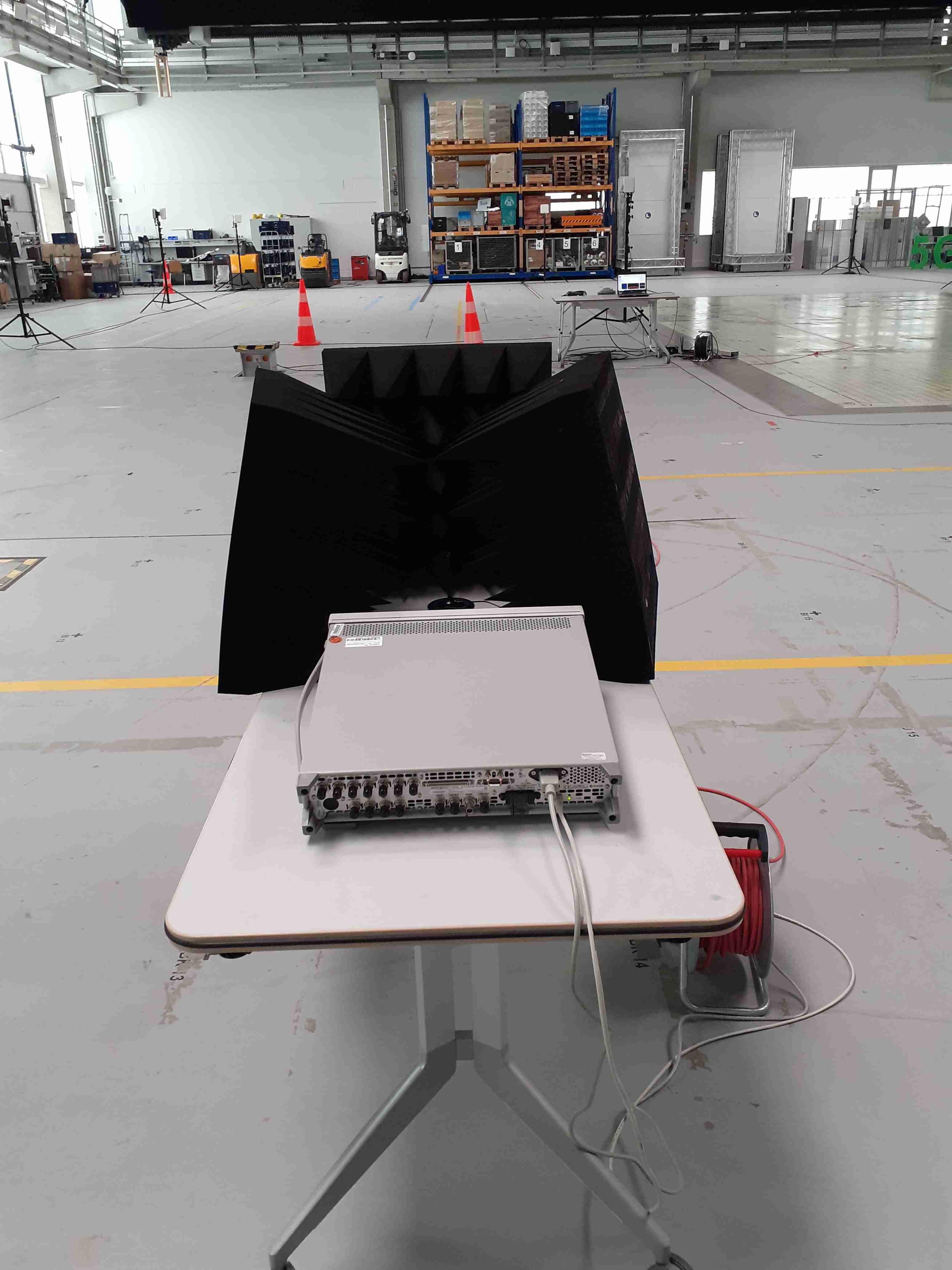}
        \vspace{-0.5cm}
    	\subcaption{Scen. 11.}
    	\label{figure_multi_path11}
    \end{minipage}
    \caption{Overview of different multipath scenarios where large black absorber walls are placed between and around the signal generator and the antenna.}
    \label{figure_multi_path}
\end{figure}

\newcommand\y{0.055}
\begin{figure*}[!t]\captionsetup[subfigure]{font=scriptsize}
    \centering
	\begin{minipage}[t]{0.072\linewidth}
        \centering
    	\includegraphics[trim=160 22 176 36, clip, width=1.0\linewidth]{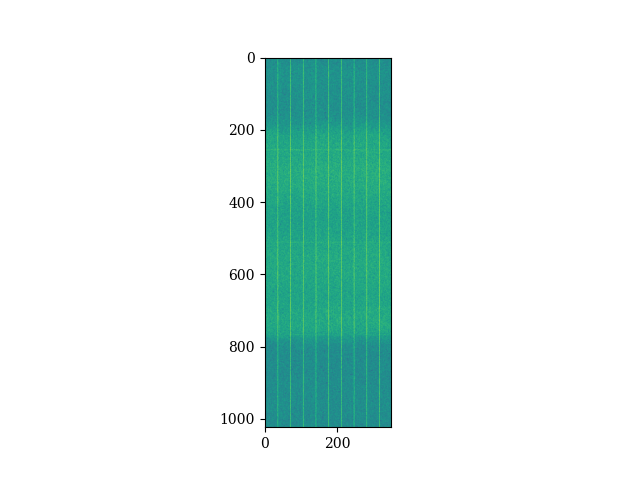}
    	\subcaption{None}
    	\label{figure_exmp_samples1}
    \end{minipage}
    \hfill
	\begin{minipage}[t]{\y\linewidth}
        \centering
    	\includegraphics[trim=189 22 176 36, clip, width=1.0\linewidth]{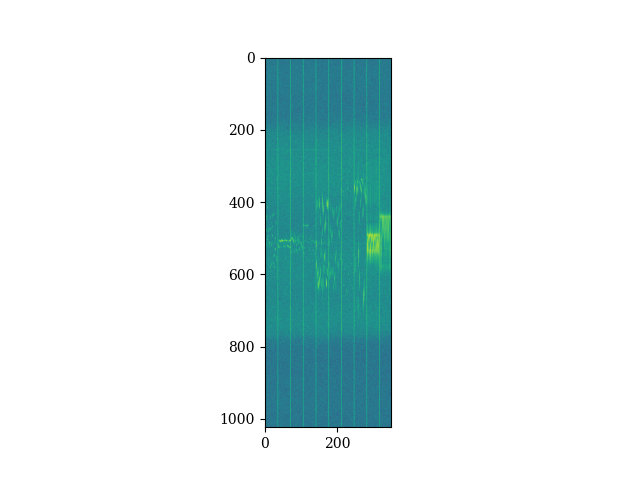}
    	\subcaption{Chirp}
    	\label{figure_exmp_samples2}
    \end{minipage}
    \hfill
	\begin{minipage}[t]{\y\linewidth}
        \centering
    	\includegraphics[trim=189 22 176 36, clip, width=1.0\linewidth]{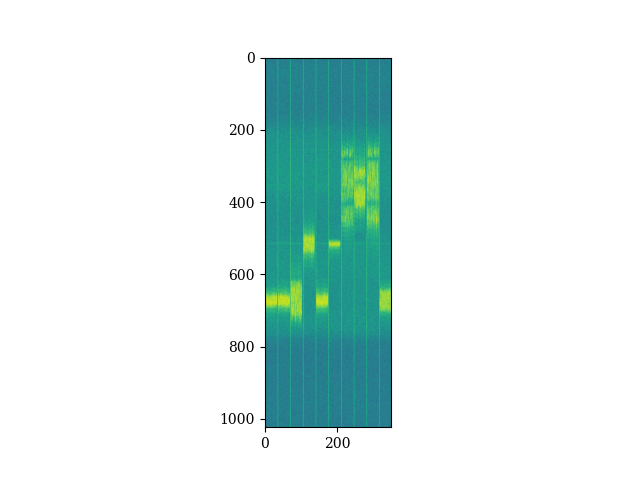}
    	\subcaption{Freq-Hopper}
    	\label{figure_exmp_samples3}
    \end{minipage}
    \hfill
	\begin{minipage}[t]{\y\linewidth}
        \centering
    	\includegraphics[trim=189 22 176 36, clip, width=1.0\linewidth]{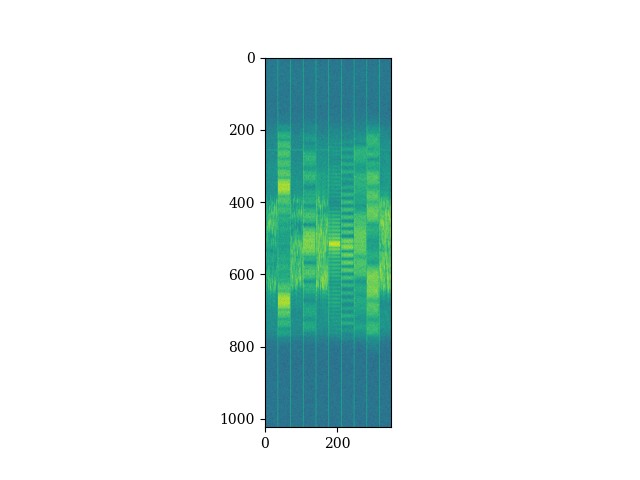}
    	\subcaption{Modulated}
    	\label{figure_exmp_samples4}
    \end{minipage}
    \hfill
	\begin{minipage}[t]{\y\linewidth}
        \centering
    	\includegraphics[trim=189 22 176 36, clip, width=1.0\linewidth]{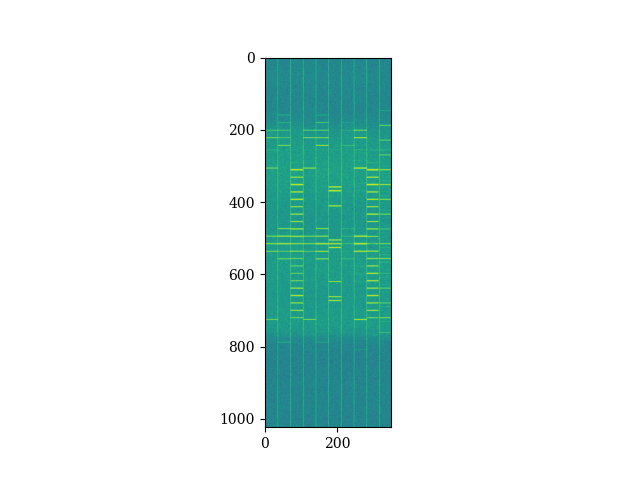}
    	\subcaption{Multitone}
    	\label{figure_exmp_samples5}
    \end{minipage}
    \hfill
	\begin{minipage}[t]{\y\linewidth}
        \centering
    	\includegraphics[trim=189 22 176 36, clip, width=1.0\linewidth]{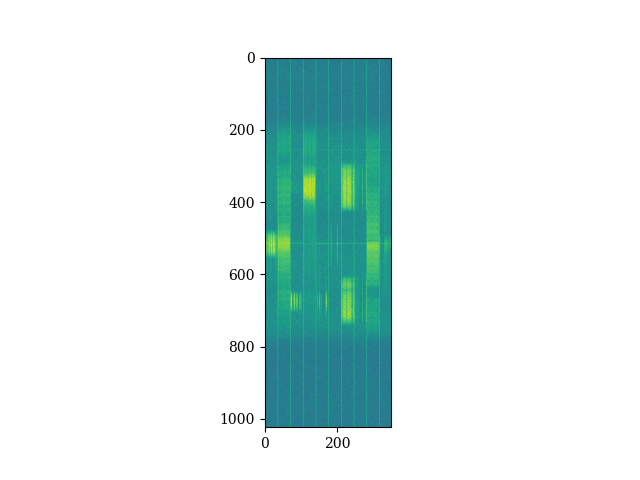}
    	\subcaption{Pulsed}
    	\label{figure_exmp_samples6}
    \end{minipage}
    \hfill
	\begin{minipage}[t]{\y\linewidth}
        \centering
    	\includegraphics[trim=189 22 176 36, clip, width=1.0\linewidth]{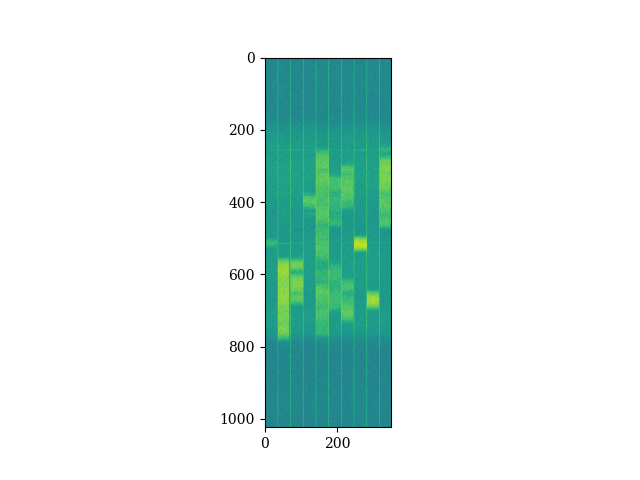}
    	\subcaption{Noise}
    	\label{figure_exmp_samples7}
    \end{minipage}
    \hfill
	\begin{minipage}[t]{\y\linewidth}
        \centering
    	\includegraphics[trim=189 22 176 36, clip, width=1.0\linewidth]{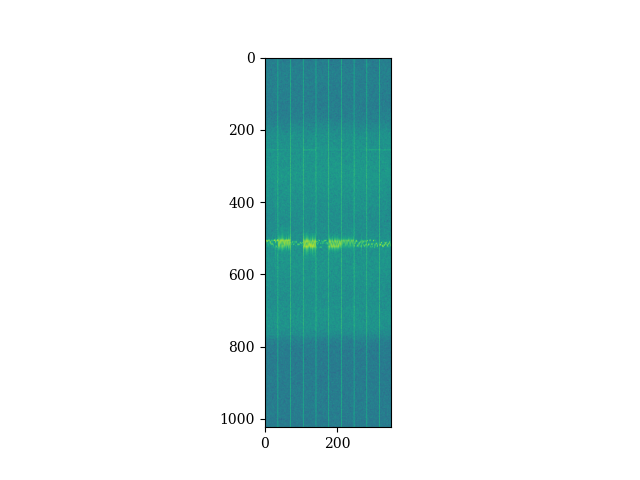}
    	\subcaption{Chirp, BW 2}
    	\label{figure_exmp_samples8}
    \end{minipage}
	\begin{minipage}[t]{\y\linewidth}
        \centering
    	\includegraphics[trim=189 22 176 36, clip, width=1.0\linewidth]{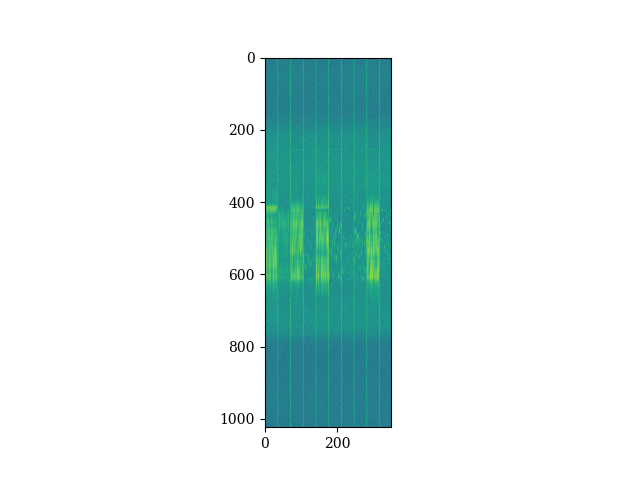}
    	\subcaption{Chirp, BW 20}
    	\label{figure_exmp_samples9}
    \end{minipage}
    \hfill
	\begin{minipage}[t]{\y\linewidth}
        \centering
    	\includegraphics[trim=189 22 176 36, clip, width=1.0\linewidth]{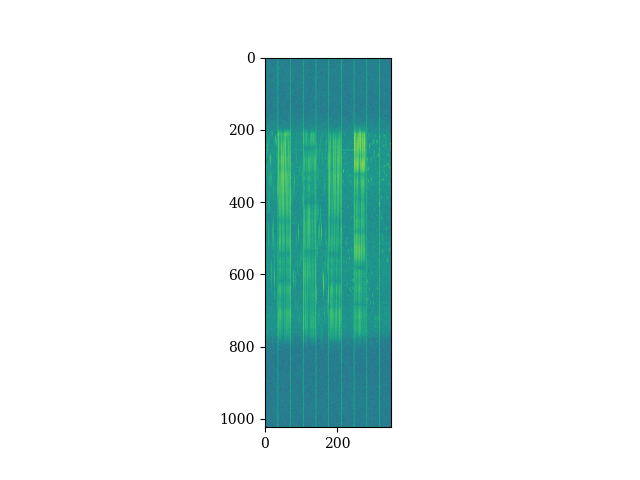}
    	\subcaption{Chirp, BW 60}
    	\label{figure_exmp_samples10}
    \end{minipage}
    \hfill
	\begin{minipage}[t]{\y\linewidth}
        \centering
    	\includegraphics[trim=189 22 176 36, clip, width=1.0\linewidth]{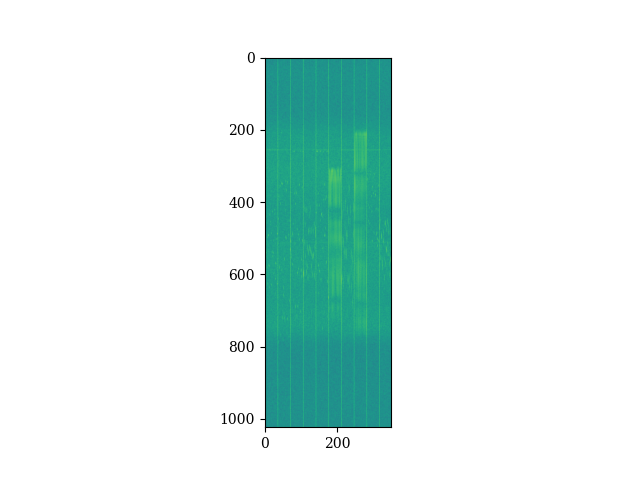}
    	\subcaption{SP -10, Chirp}
    	\label{figure_exmp_samples11}
    \end{minipage}
    \hfill
	\begin{minipage}[t]{\y\linewidth}
        \centering
    	\includegraphics[trim=189 22 176 36, clip, width=1.0\linewidth]{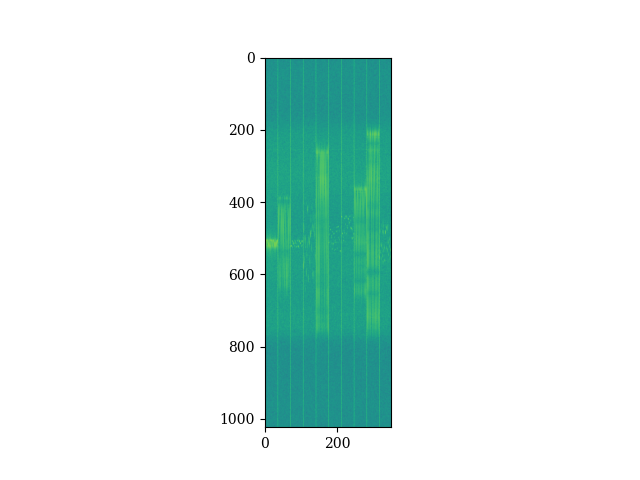}
    	\subcaption{SP 4, Chirp}
    	\label{figure_exmp_samples12}
    \end{minipage}
    \hfill
	\begin{minipage}[t]{\y\linewidth}
        \centering
    	\includegraphics[trim=189 22 176 36, clip, width=1.0\linewidth]{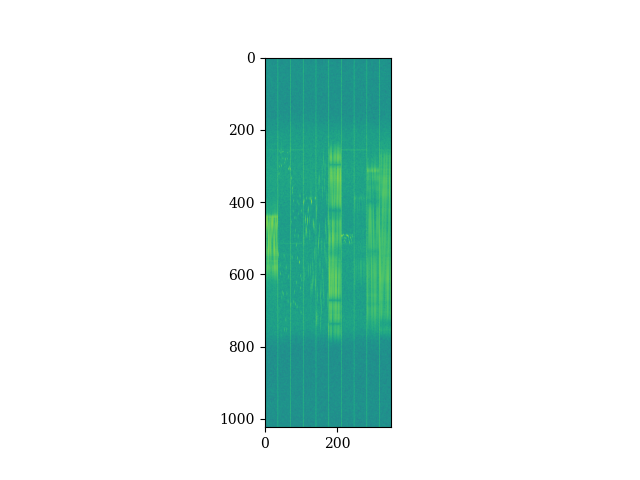}
    	\subcaption{SP 10, Chirp}
    	\label{figure_exmp_samples13}
    \end{minipage}
    \hfill
	\begin{minipage}[t]{\y\linewidth}
        \centering
    	\includegraphics[trim=189 22 176 36, clip, width=1.0\linewidth]{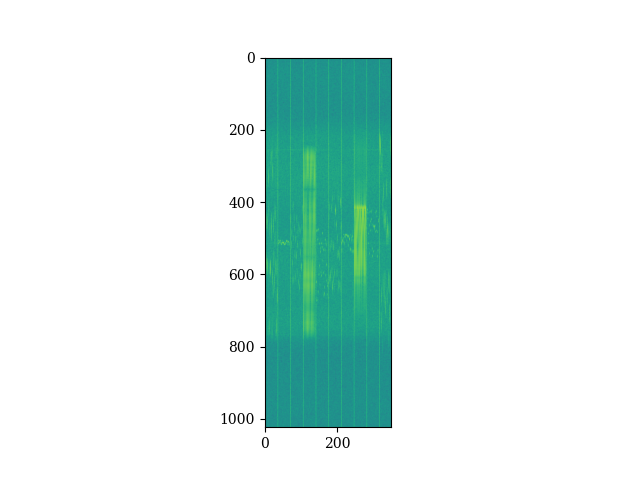}
    	\subcaption{Chirp, scen. 1}
    	\label{figure_exmp_samples14}
    \end{minipage}
    \hfill
	\begin{minipage}[t]{\y\linewidth}
        \centering
    	\includegraphics[trim=189 22 176 36, clip, width=1.0\linewidth]{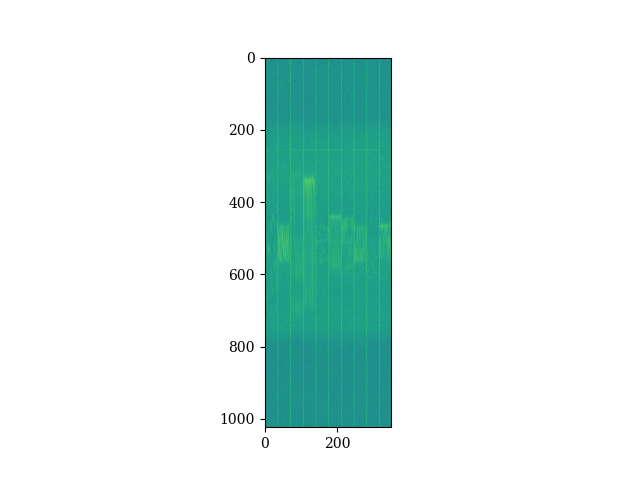}
    	\subcaption{Chirp, scen. 7}
    	\label{figure_exmp_samples15}
    \end{minipage}
    \hfill
	\begin{minipage}[t]{\y\linewidth}
        \centering
    	\includegraphics[trim=189 22 176 36, clip, width=1.0\linewidth]{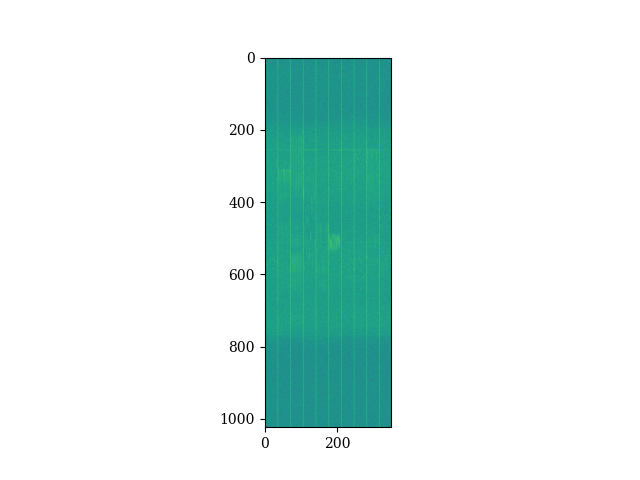}
    	\subcaption{Chirp, scen. 8}
    	\label{figure_exmp_samples16}
    \end{minipage}
    \caption{Exemplary spectrogram samples of the non-interference class (a) and all six interference types (b to g), a signal with chirp interference with different bandwidths (BW) (h to j) and signal powers (SP) (k to m), and a chirp interference from the scenario 1 (open environment), scenario 7 (see Figure~\ref{figure_multi_path6}) and scenario 8 (see Figure~\ref{figure_multi_path7}). We concatenate ten spectrograms, each with $\text{N}_t=34$, separated by a thin line, to visually emphasize the variability of the interference.}
    \label{figure_exmp_samples}
\end{figure*}

\setlength{\intextsep}{6pt}
\setlength{\columnsep}{12pt}
\begin{wraptable}{R}{4.1cm}
    \begin{minipage}[b]{1.0\linewidth}
    \begin{center}
        \setlength{\tabcolsep}{2.0pt}
        \vspace{-0.2cm}
        \caption{Overview of the amount of snapshots w.r.t.~various BWs [in MHz].}
        \label{table_bandwidth}
        \scriptsize \begin{tabular}{ p{0.5cm} | p{0.5cm} | p{0.5cm} | p{0.5cm} | p{0.5cm} | p{0.5cm} }
        \multicolumn{2}{c|}{\textbf{Chirp}} & \multicolumn{2}{c|}{\textbf{FreqHopper}} & \multicolumn{2}{c}{\textbf{Pulsed}} \\ 
        \multicolumn{1}{c|}{\textbf{BW}} & \multicolumn{1}{c|}{\textbf{\#}} & \multicolumn{1}{c|}{\textbf{BW}} & \multicolumn{1}{c|}{\textbf{\#}} & \multicolumn{1}{c|}{\textbf{BW}} & \multicolumn{1}{c}{\textbf{\#}} \\ \hline
        \multicolumn{1}{r|}{2.0} & \multicolumn{1}{r|}{2,584} & \multicolumn{1}{r|}{0.1} & \multicolumn{1}{r|}{36} & \multicolumn{1}{r|}{0.2} & \multicolumn{1}{r}{84} \\
        \multicolumn{1}{r|}{5.0} & \multicolumn{1}{r|}{2,584} & \multicolumn{1}{r|}{0.5} & \multicolumn{1}{r|}{36} & \multicolumn{1}{r|}{1.0} & \multicolumn{1}{r}{148} \\
        \multicolumn{1}{r|}{10.0} & \multicolumn{1}{r|}{2,584} & \multicolumn{1}{r|}{1.0} & \multicolumn{1}{r|}{36} & \multicolumn{1}{r|}{2.5} & \multicolumn{1}{r}{252} \\
        \multicolumn{1}{r|}{15.0} & \multicolumn{1}{r|}{2,584} & \multicolumn{1}{r|}{2.0} & \multicolumn{1}{r|}{36} & \multicolumn{1}{r|}{4.0} & \multicolumn{1}{r}{84} \\
        \multicolumn{1}{r|}{20.0} & \multicolumn{1}{r|}{2,584} & \multicolumn{1}{r|}{2.5} & \multicolumn{1}{r|}{108} & \multicolumn{1}{r|}{5.0} & \multicolumn{1}{r}{252} \\
        \multicolumn{1}{r|}{25.0} & \multicolumn{1}{r|}{2,768} & \multicolumn{1}{r|}{4.0} & \multicolumn{1}{r|}{36} & \multicolumn{1}{r|}{10.0} & \multicolumn{1}{r}{348} \\
        \multicolumn{1}{r|}{30.0} & \multicolumn{1}{r|}{2,584} & \multicolumn{1}{r|}{5.0} & \multicolumn{1}{r|}{252} & \multicolumn{1}{r|}{35.0} & \multicolumn{1}{r}{84} \\
        \multicolumn{1}{r|}{35.0} & \multicolumn{1}{r|}{2,584} & \multicolumn{1}{r|}{10.0} & \multicolumn{1}{r|}{252} & \multicolumn{1}{r|}{50.0} & \multicolumn{1}{r}{84} \\
        \multicolumn{1}{r|}{40.0} & \multicolumn{1}{r|}{2,584} & \multicolumn{1}{r|}{20.0} & \multicolumn{1}{r|}{144} & \multicolumn{1}{r}{} & \multicolumn{1}{r}{} \\
        \multicolumn{1}{r|}{50.0} & \multicolumn{1}{r|}{2,584} & \multicolumn{1}{r|}{25.0} & \multicolumn{1}{r|}{100} & \multicolumn{1}{r}{} & \multicolumn{1}{r}{} \\
        \multicolumn{1}{r|}{60.0} & \multicolumn{1}{r|}{2,584} & \multicolumn{1}{r|}{35.0} & \multicolumn{1}{r|}{36} & \multicolumn{1}{r}{} & \multicolumn{1}{r}{} \\
        \multicolumn{1}{r}{} & \multicolumn{1}{r|}{} & \multicolumn{1}{r|}{50.0} & \multicolumn{1}{r|}{36} & \multicolumn{1}{r}{} & \multicolumn{1}{r}{} \\
        \end{tabular}
    \end{center}
    \end{minipage}
\end{wraptable}
The primary aim is to develop ML models characterized by resilience against an array of jammer types, interference profiles, antenna variations, environmental fluctuations, shifts in location, and disparate receiver stations. Consequently, we propose the utilization of a GNSS-based dataset to facilitate the analysis of model robustness, accompanied by detailed experiment definitions. The data recording setup is structured as follows: A spacious indoor hall measuring $1,320\,m^2$ is designated as the recording environment to enable controlled data collection encompassing multipath effects. A receiver antenna, i.e., a patch antenna array consisting of four identical coaxial-fed square patch elements, is situated on one end of the hall and an MXG vector-signal generator on the opposite end (refer to Figure~\ref{figure_introduction}). The signal generator is designed to produce high-quality radio frequency (RF) signals, characterized by excellent signal purity, high output power, extensive frequency coverage, and adaptable modulation capabilities. Subsequently, the antenna's signals are recorded as snapshots that include various interferences from the signal generator. A substantial dataset is recorded under diverse setups, encompassing scenarios within an unoccupied environment and configurations featuring absorber walls interposed between the antenna and the generator (refer to Figure~\ref{figure_multi_path}). This setup facilitates the recording of distinct multipath effects, spanning from minor manifestations (Figure~\ref{figure_multi_path2} and Figure~\ref{figure_multi_path3}) to pronounced effects (Figure~\ref{figure_multi_path5}), and instances of significant absorption (Figure~\ref{figure_multi_path8}). We have a data recording at $1.57542\,\text{GHz}$ with a BW of $100\,\text{MHz}$ quadrature sampled with a duration of $10\,\mu s$. We create (non-overlapping) spectrograms using an Fast Fourier transform (FFT) with window size 1,024. Each spectrogram comprises 1,024 timesteps and has dimensions of $1,024 \times \text{N}_{\text{t}}$, where $\text{N}_{\text{t}}$ is the snapshot length that we set to a maximum of 34. Figure~\ref{figure_exmp_samples} illustrates a variety of aggregatd spectrograms of the snapshots, with each image displaying 10 randomly selected samples. Figure~\ref{figure_exmp_samples1} depicts snapshots devoid of interference. To introduce interference, six distinct types are generated, namely Chirp, FreqHopper, Modulated, Multitone, Pulsed, and Noise (refer to Figure~\ref{figure_exmp_samples2} to Figure~\ref{figure_exmp_samples7}). For further information on common interference typologies, consult \cite{brieger_ion_gnss}. The primary goal is to detect and accurately classify these interferences. Additionally, the jammer types necessitate characterization, including the BW and signal power. Thus, we record various BWs, spanning from $0.1\,\text{MHz}$ to $60\,\text{MHz}$ (observe a wider interference spectrum across multiple channels in Figure~\ref{figure_exmp_samples8} to Figure~\ref{figure_exmp_samples10} for Chirp), and signal power, ranging from -10 to 10, set at the MXG (note heightened intensity from Figure~\ref{figure_exmp_samples11} to Figure~\ref{figure_exmp_samples13}). Enhanced multipath effects correspond to diminished interference intensity (refer to Figure~\ref{figure_exmp_samples15} for scenario 7).

An overview of all recorded scenarios, interferences, and the corresponding sample counts is presented in Table~\ref{table_dataset}. Scenario 1 is characterized by the absence of absorber walls. Initially, we captured data from 16 distinct positions of the signal generator across the hall, maintaining a signal power of 10. Subsequently, the generator was stationed at a fixed position while we recorded data encompassing various signal powers, either $6\,\text{dBm}$, $8\,\text{dBm}$, or $10\,\text{dBm}$ across all interference types or spanning from $-10\,\text{dBm}$ to $10\,\text{dBm}$ with a step increment of $2\,\text{dBm}$ for a subset of interferences. Additionally, interferences were recorded at 30 generator positions situated on the gallery of the top of the hall, with a signal power of $6\,\text{dBm}$. For scenarios 2 to 11, a variable number of absorber walls were introduced as depicted in Figure~\ref{figure_multi_path}. In total, the dataset comprises 42,592 samples, of which 576 samples are devoid of any interferences. Table~\ref{table_bandwidth} provides a summary of the sample counts corresponding to different BWs for three interference types.

\begin{table}[t!]
\begin{center}
\setlength{\tabcolsep}{1.6pt}
    \vspace{0.2cm}
    \caption{Overview of the dataset recorded with an antenna recording platform and a signal generator (MXG) in an open environment with no multipath effects (scenario 1) or with multipath effects from absorber walls (AW) (scenario 2 to 11).}
    \label{table_dataset}
    \scriptsize \begin{tabular}{ p{0.5cm} | p{0.5cm} | p{0.5cm} | p{0.5cm} | p{0.5cm} | p{0.5cm} | p{0.5cm} | p{0.5cm} | p{0.5cm} | p{0.5cm} | p{0.5cm} | p{0.5cm} | p{0.5cm} }
    \multicolumn{1}{c|}{} & \multicolumn{7}{c|}{\textbf{Interference}} & \multicolumn{2}{c|}{\textbf{Size}} & \multicolumn{1}{c|}{} & \multicolumn{1}{c|}{} & \multicolumn{1}{c}{} \\
    \multicolumn{1}{c|}{\rot{90}{\textbf{Scenario}}} & \multicolumn{1}{c}{\rot{90}{None (0)}} & \multicolumn{1}{c}{\rot{90}{Noise (1)}} & \multicolumn{1}{c}{\rot{90}{Chirp (2)}} & \multicolumn{1}{c}{\rot{90}{FreqHopper (3)}} & \multicolumn{1}{c}{\rot{90}{Modulated (4)}} & \multicolumn{1}{c}{\rot{90}{Multitone (5)}} & \multicolumn{1}{c|}{\rot{90}{Pulsed (6)}} & \multicolumn{1}{c}{\rot{90}{non-interference}} & \multicolumn{1}{c|}{\rot{90}{interference}} & \multicolumn{1}{c|}{\rot{90}{\textbf{Signal power [dBm]}}} & \multicolumn{1}{c|}{\rot{90}{\textbf{Figure}}} & \multicolumn{1}{c}{\textbf{Note}} \\ \hline
    \multicolumn{1}{c|}{1} & \multicolumn{1}{c}{x} & \multicolumn{1}{c}{x} & \multicolumn{1}{c}{x} & \multicolumn{1}{c}{x} & \multicolumn{1}{c}{x} & \multicolumn{1}{c}{x} & \multicolumn{1}{c|}{x} & \multicolumn{1}{r}{576} & \multicolumn{1}{r|}{240} & \multicolumn{1}{c|}{10} & \multicolumn{1}{c|}{-} & \multicolumn{1}{l}{no multipath, 16 positions} \\
    \multicolumn{1}{c|}{1} & \multicolumn{1}{c}{x} & \multicolumn{1}{c}{x} & \multicolumn{1}{c}{x} & \multicolumn{1}{c}{x} & \multicolumn{1}{c}{x} & \multicolumn{1}{c}{x} & \multicolumn{1}{c|}{x} & \multicolumn{1}{r}{576} & \multicolumn{1}{r|}{3,720} & \multicolumn{1}{c|}{[6, 8, 10]} & \multicolumn{1}{c|}{-} & \multicolumn{1}{l}{no multipath, fixed position} \\
    \multicolumn{1}{c|}{1} & \multicolumn{1}{c}{x} & \multicolumn{1}{c}{x} & \multicolumn{1}{c}{x} & \multicolumn{1}{c}{} & \multicolumn{1}{c}{} & \multicolumn{1}{c}{x} & \multicolumn{1}{c|}{} & \multicolumn{1}{r}{576} & \multicolumn{1}{r|}{12,360} & \multicolumn{1}{c|}{[-10, $\ldots$, 10]} & \multicolumn{1}{c|}{-} & \multicolumn{1}{l}{no multipath, fixed position} \\
    \multicolumn{1}{c|}{1} & \multicolumn{1}{c}{x} & \multicolumn{1}{c}{x} & \multicolumn{1}{c}{x} & \multicolumn{1}{c}{} & \multicolumn{1}{c}{} & \multicolumn{1}{c}{} & \multicolumn{1}{c|}{} & \multicolumn{1}{r}{576} & \multicolumn{1}{r|}{384} & \multicolumn{1}{c|}{6} & \multicolumn{1}{c|}{-} & \multicolumn{1}{l}{on gallery, 30 positions} \\
    \multicolumn{1}{c|}{2} & \multicolumn{1}{c}{x} & \multicolumn{1}{c}{} & \multicolumn{1}{c}{x} & \multicolumn{1}{c}{} & \multicolumn{1}{c}{} & \multicolumn{1}{c}{x} & \multicolumn{1}{c|}{} & \multicolumn{1}{r}{576} & \multicolumn{1}{r|}{11,520} & \multicolumn{1}{c|}{[2, $\ldots$, 10]} & \multicolumn{1}{l|}{\ref{figure_multi_path2}} & \multicolumn{1}{l}{1 AW to MXG} \\
    \multicolumn{1}{c|}{3} & \multicolumn{1}{c}{x} & \multicolumn{1}{c}{x} & \multicolumn{1}{c}{x} & \multicolumn{1}{c}{} & \multicolumn{1}{c}{} & \multicolumn{1}{c}{x} & \multicolumn{1}{c|}{} & \multicolumn{1}{r}{576} & \multicolumn{1}{r|}{1,648} & \multicolumn{1}{c|}{[6, 10]} & \multicolumn{1}{l|}{\ref{figure_multi_path3}} & \multicolumn{1}{l}{1 AW to antenna} \\
    \multicolumn{1}{c|}{4} & \multicolumn{1}{c}{x} & \multicolumn{1}{c}{x} & \multicolumn{1}{c}{x} & \multicolumn{1}{c}{} & \multicolumn{1}{c}{} & \multicolumn{1}{c}{x} & \multicolumn{1}{c|}{} & \multicolumn{1}{r}{576} & \multicolumn{1}{r|}{1,648} & \multicolumn{1}{c|}{[6, 10]} & \multicolumn{1}{l|}{\ref{figure_multi_path4}} & \multicolumn{1}{l}{3 AW + small AW to MXG} \\
    \multicolumn{1}{c|}{5} & \multicolumn{1}{c}{x} & \multicolumn{1}{c}{x} & \multicolumn{1}{c}{x} & \multicolumn{1}{c}{} & \multicolumn{1}{c}{} & \multicolumn{1}{c}{} & \multicolumn{1}{c|}{} & \multicolumn{1}{r}{576} & \multicolumn{1}{r|}{1,552} & \multicolumn{1}{c|}{[6, 10]} & \multicolumn{1}{l|}{\ref{figure_multi_path5}} & \multicolumn{1}{l}{3 AW to MXG} \\
    \multicolumn{1}{c|}{6} & \multicolumn{1}{c}{x} & \multicolumn{1}{c}{x} & \multicolumn{1}{c}{x} & \multicolumn{1}{c}{} & \multicolumn{1}{c}{} & \multicolumn{1}{c}{x} & \multicolumn{1}{c|}{} & \multicolumn{1}{r}{576} & \multicolumn{1}{r|}{1,648} & \multicolumn{1}{c|}{[6, 10]} & \multicolumn{1}{l|}{\ref{figure_multi_path6}} & \multicolumn{1}{l}{4 AW to antenna} \\
    \multicolumn{1}{c|}{7} & \multicolumn{1}{c}{x} & \multicolumn{1}{c}{} & \multicolumn{1}{c}{x} & \multicolumn{1}{c}{} & \multicolumn{1}{c}{} & \multicolumn{1}{c}{} & \multicolumn{1}{c|}{} & \multicolumn{1}{r}{576} & \multicolumn{1}{r|}{704} & \multicolumn{1}{c|}{[6, 10]} & \multicolumn{1}{l|}{\ref{figure_multi_path7}} & \multicolumn{1}{l}{4 AW to MXG} \\
    \multicolumn{1}{c|}{8} & \multicolumn{1}{c}{x} & \multicolumn{1}{c}{x} & \multicolumn{1}{c}{x} & \multicolumn{1}{c}{} & \multicolumn{1}{c}{} & \multicolumn{1}{c}{x} & \multicolumn{1}{c|}{} & \multicolumn{1}{r}{576} & \multicolumn{1}{r|}{1,648} & \multicolumn{1}{c|}{[6, 10]} & \multicolumn{1}{l|}{\ref{figure_multi_path8}} & \multicolumn{1}{l}{4 AW + small AW to MXG} \\
    \multicolumn{1}{c|}{9} & \multicolumn{1}{c}{x} & \multicolumn{1}{c}{x} & \multicolumn{1}{c}{x} & \multicolumn{1}{c}{} & \multicolumn{1}{c}{} & \multicolumn{1}{c}{x} & \multicolumn{1}{c|}{} & \multicolumn{1}{r}{576} & \multicolumn{1}{r|}{1,648} & \multicolumn{1}{c|}{[6, 10]} & \multicolumn{1}{l|}{\ref{figure_multi_path9}} & \multicolumn{1}{l}{4 AW as corridor} \\
    \multicolumn{1}{c|}{10} & \multicolumn{1}{c}{x} & \multicolumn{1}{c}{x} & \multicolumn{1}{c}{x} & \multicolumn{1}{c}{} & \multicolumn{1}{c}{} & \multicolumn{1}{c}{x} & \multicolumn{1}{c|}{} & \multicolumn{1}{r}{576} & \multicolumn{1}{r|}{1,648} & \multicolumn{1}{c|}{[6, 10]} & \multicolumn{1}{l|}{\ref{figure_multi_path10}} & \multicolumn{1}{l}{4 AW to MXG} \\
    \multicolumn{1}{c|}{11} & \multicolumn{1}{c}{x} & \multicolumn{1}{c}{x} & \multicolumn{1}{c}{x} & \multicolumn{1}{c}{} & \multicolumn{1}{c}{} & \multicolumn{1}{c}{x} & \multicolumn{1}{c|}{} & \multicolumn{1}{r}{576} & \multicolumn{1}{r|}{1,648} & \multicolumn{1}{c|}{[6, 10]} & \multicolumn{1}{l|}{\ref{figure_multi_path11}} & \multicolumn{1}{l}{3 small AW around MXG} \\
    \end{tabular}
\end{center}
\end{table}

\subsection{Definition of Experiments}
\label{label_experiments_definition}

In this section, we articulate the experiments feasible with the proposed datasets, aimed at assessing the model's robustness. These experiments encompass: (1) Classification of seven distinct categories, comprising six types of interference and a single category denoting non-interference. (2) Characterization of interference by categorizing the energy level; the signal power with noise. (3) Characterization of interference by categorizing the BW. (4) Evaluation of diverse train-test splits: a dependent split, where both the training and testing sets possess identical interference characteristics (i.e., BW), and an independent split, where such characteristics differ. This facilitates the assessment of model generalizability. (5) Classification of antenna regions to anticipate the angle of interference. (6) Utilization of the recorded dataset for interference localization, entailing prediction of the signal generator's position (relative position between the MXG and the antenna based on labels). Accordingly, the model is trained to predict positions within the open hall (16 positions), on the gallery (30 positions), or a combination of both (46 positions). (7) Cross-validation across 12 distinct training and testing scenarios, enabling analysis of environmental variability and robustness against multipath effects, particularly increased absorption. (8) Investigation of snapshot length, initially set at a maximum of 34, which is systematically reduced to a minimum of 1. This exploration aims to determine the minimal data requirement for achieving high classification accuracies across each task.

%% file: 04method.tex
\section{Methodology}
\label{label_method}

\textbf{Pipeline.} For the purpose of evaluation, we employ the subsequent pipeline. The dimensions of the model inputs are $1,024 \times \text{N}_{\text{t}}$, where $\text{N}_{\text{t}}$ represents the snapshot length. Initially, we train with a sample length of $\text{N}_{\text{t}} = 34$, which we subsequently reduce to $\text{N}_{\text{t}} = 1$ to explore the minimum requisite of data information. Each sample comprises three image channels, and we apply min-max normalization with values of -182.77 and -17.12. Considering the notable performance of ResNet18\cite{he_zhang} across interference classification tasks \cite{brieger_ion_gnss,ott_heublein,raichur_heublein}, we employ ResNet18 for conducting our experiments. Initially, we pre-train ResNet18 on the ImageNet dataset. We then remove the final fully connected (FC) layer and append, for each task, a distinct FC layer of dimensions $512 \times \text{N}_{\text{c}}$, where $\text{N}_{\text{c}}$ denotes the number of classes. Specifically, the interference classification task involves $\text{N}_{\text{c}} = 7$ classes, the antenna area task comprises $\text{N}_{\text{c}} = 4$ classes, and the regression task entails the prediction of single values for the signal power, BW, and direction angle. To address both classification and regression objectives, we combine cross-entropy loss for classification and root mean squared error (RMSE) loss for regression. This is achieved by aggregating up to three loss functions, each weighted equally at 1.

\textbf{Uncertainty Quantification.} The primary objective is to enhance model robustness concerning specific environmental factors through the utilization of Bayesian inference. This involves computing the posterior distribution $p(\theta|D)$, which represents the neural network (NN) weights, given the training dataset $D$ and the model parameters $\theta$. However, due to the typical intractability of computing the posterior, a (local) approximation is commonly employed \cite{klass_lorenz_strl}. We utilize Deep Ensembles, comprising a committee of $M$ individual NNs initialized with distinct seeds, where the initialization serves as the sole source of stochasticity in the model parameters. The results are derived by aggregating predictions from $M=10$ independently trained NNs. Next, we decompose uncertainty into \textit{aleatoric} (represents stochasticity inherent within the data) and \textit{epistemic} (can be diminished with an increasing amount of observations) uncertainty by the methodology proposed by Kwon et al.~\cite{kwon_won} based on the variability of the softmax output \cite{klass_lorenz_strl}.

\textbf{Architecture Benchmark.} We evaluate a total of 129 vision models on the classification tasks previously described -- namely, classifying interference type, antenna area, signal power, and multipath scenario -- as well as on the regression tasks of predicting signal power and bandwidth. To achieve this, we employ 19 well-established time-series models proposed by timeseriesAI (tsai)~\cite{tsai} and 110 PyTorch image models from Hugging Face~\cite{rw2019timm}. Given that some models require specific input dimensions, we downsample the snapshot input from 1,024 to 512 and interpolate it accordingly. Our results indicate that the ResNet18 model consistently performs among the best across all tasks. Consequently, we select ResNet18 for further in-depth analysis.

%% file: 05evaluation.tex
\section{Evaluation}
\label{label_evaluation}

\begin{table}[t!]
\begin{center}
\setlength{\tabcolsep}{1.2pt}
    \caption{Overview of evaluation results for the combination of the interference type classification task (mean and standard deviation accuracy in \% and weighted F2-score), a second classification task, and a regression task. $^{\text{1}}$Only sub-interferences that are overlapping with the gallery dataset.}
    \label{table_all_results}
    \scriptsize \begin{tabular}{ p{0.5cm} | p{0.5cm} | p{0.5cm} | p{0.5cm} | p{0.5cm} | p{0.5cm} }
    \multicolumn{1}{c|}{} & \multicolumn{1}{c|}{} & \multicolumn{2}{c|}{\textbf{Interference type}} & \multicolumn{1}{c|}{\textbf{Task 2}} & \multicolumn{1}{c}{\textbf{Task 3}} \\
    \multicolumn{1}{c|}{\textbf{Task 1}} & \multicolumn{1}{c|}{\textbf{Task 2 or Task 3}} & \multicolumn{1}{c|}{\textbf{Accuracy}} & \multicolumn{1}{c|}{\textbf{F2-score}} & \multicolumn{1}{c|}{\textbf{Accuracy}} & \multicolumn{1}{c}{\textbf{Error [$\cdot$]}} \\ \hline
    \multicolumn{1}{l|}{Intf. type (dep.)} & \multicolumn{1}{l|}{Signal power} & \multicolumn{1}{r|}{99.88\,\tiny{$\pm$\,0.04}} & \multicolumn{1}{r|}{99.88\,\tiny{$\pm$\,0.04}} & \multicolumn{1}{r|}{86.42\,\tiny{$\pm$\,1.18}} & \multicolumn{1}{c}{-} \\
    \multicolumn{1}{l|}{Intf. type (dep.)} & \multicolumn{1}{l|}{Antenna area} & \multicolumn{1}{r|}{99.88\,\tiny{$\pm$\,0.04}} & \multicolumn{1}{r|}{99.88\,\tiny{$\pm$\,0.04}} & \multicolumn{1}{r|}{98.26\,\tiny{$\pm$\,1.87}} & \multicolumn{1}{c}{-} \\
    \multicolumn{1}{l|}{Intf. type (indep.)} & \multicolumn{1}{l|}{Signal power} & \multicolumn{1}{r|}{96.15\,\tiny{$\pm$\,0.44}} & \multicolumn{1}{r|}{95.71\,\tiny{$\pm$\,0.44}} & \multicolumn{1}{r|}{55.96\,\tiny{$\pm$\,0.92}} & \multicolumn{1}{c}{-} \\
    \multicolumn{1}{l|}{Intf. type (indep.)} & \multicolumn{1}{l|}{Antenna area} & \multicolumn{1}{r|}{96.15\,\tiny{$\pm$\,0.44}} & \multicolumn{1}{r|}{95.71\,\tiny{$\pm$\,0.44}} & \multicolumn{1}{r|}{98.62\,\tiny{$\pm$\,0.37}} & \multicolumn{1}{c}{-} \\
    \multicolumn{1}{l|}{Intf. type (dep.)} & \multicolumn{1}{l|}{Signal power} & \multicolumn{1}{r|}{94.44\,\tiny{$\pm$\,7.99}} & 
    \multicolumn{1}{r|}{93.92\,\tiny{$\pm$\,9.00}} & \multicolumn{1}{c|}{-} & \multicolumn{1}{l}{0.81\,\scriptsize{dBm}} \\
    \multicolumn{1}{c|}{-} & \multicolumn{1}{l|}{Bandwidth} & \multicolumn{1}{c|}{-} & \multicolumn{1}{c|}{-} & \multicolumn{1}{c|}{-} & \multicolumn{1}{l}{0.87\,\scriptsize{MHz}} \\
    \multicolumn{1}{l|}{Intf. type (rand.)} & \multicolumn{1}{l|}{Local. (area)} & \multicolumn{1}{r|}{97.81\,\tiny{$\pm$\,0.61}} & \multicolumn{1}{r|}{97.75\,\tiny{$\pm$\,0.62}} & \multicolumn{1}{r|}{86.38\,\tiny{$\pm$\,1.71}} & \multicolumn{1}{c}{-} \\
    \multicolumn{1}{l|}{Intf. type (rand.)} & \multicolumn{1}{l|}{Local. (area$^{\text{1}}$)} & \multicolumn{1}{r|}{100.00\,\tiny{$\pm$\,0.00}} & \multicolumn{1}{r|}{100.00\,\tiny{$\pm$\,0.00}} & \multicolumn{1}{r|}{97.57\,\tiny{$\pm$\,0.87}} & \multicolumn{1}{c}{-} \\
    \multicolumn{1}{l|}{Intf. type (rand.)} & \multicolumn{1}{l|}{Local. (gallery)} & \multicolumn{1}{r|}{100.00\,\tiny{$\pm$\,0.00}} & \multicolumn{1}{r|}{100.00\,\tiny{$\pm$\,0.00}} & \multicolumn{1}{r|}{91.93\,\tiny{$\pm$\,2.11}} & \multicolumn{1}{c}{-} \\
    \multicolumn{1}{l|}{Intf. type (rand.)} & \multicolumn{1}{l|}{Local. (area$^{\text{1}}$+gal.)} & \multicolumn{1}{r|}{99.95\,\tiny{$\pm$\,0.14}} & \multicolumn{1}{r|}{99.95\,\tiny{$\pm$\,0.14}} & \multicolumn{1}{r|}{86.49\,\tiny{$\pm$\,1.90}} & \multicolumn{1}{c}{-} \\
    \multicolumn{1}{l|}{Intf. type (rand.)} & \multicolumn{1}{l|}{Multipath} & \multicolumn{1}{r|}{99.89\,\tiny{$\pm$\,0.08}} & \multicolumn{1}{r|}{99.89\,\tiny{$\pm$\,0.08}} & \multicolumn{1}{r|}{82.14\,\tiny{$\pm$\,0.45}} & \multicolumn{1}{c}{-} \\
    \end{tabular}
\end{center}
\end{table}

For all experiments, we utilize the same training setup.\footnote{We use Nvidia Tesla V100-SXM2 GPUs with 32 GB VRAM equipped with Core Xeon CPUs and 192\,GB RAM, the vanilla SGD optimizer with a multi-step learning rate of $0.01$, decay of $0.0005$, and train the models for 200 epochs. We train each model 10 times and present the mean and standard variance accuracy (ratio between correctly classified samples and total number of samples) and weighted F2-score ($F_2 = \frac{5 \cdot \text{precision} \cdot \text{recall}}{4 \cdot \text{precision} + \text{recall}}$) for classification tasks, as well as the mean absolute error (MAE) for regression tasks.} Table~\ref{table_all_results} provides a summary of all evaluation results.

\begin{figure}[!t]
    \centering
	\begin{minipage}[t]{1.0\linewidth}
        \centering
        \includegraphics[trim=10 14 12 34, clip, width=1.0\linewidth]{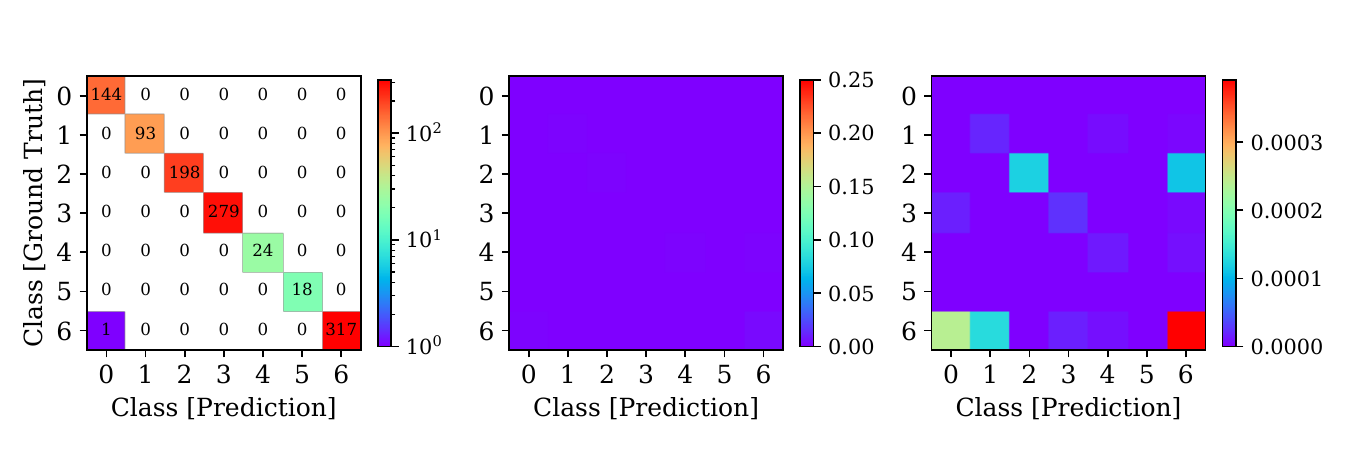}
        \subcaption{Classification of interference type.}
        \label{figure_classification_results1}
    \end{minipage}
	\begin{minipage}[t]{1.0\linewidth}
        \centering
        \includegraphics[trim=10 14 12 34, clip, width=1.0\linewidth]{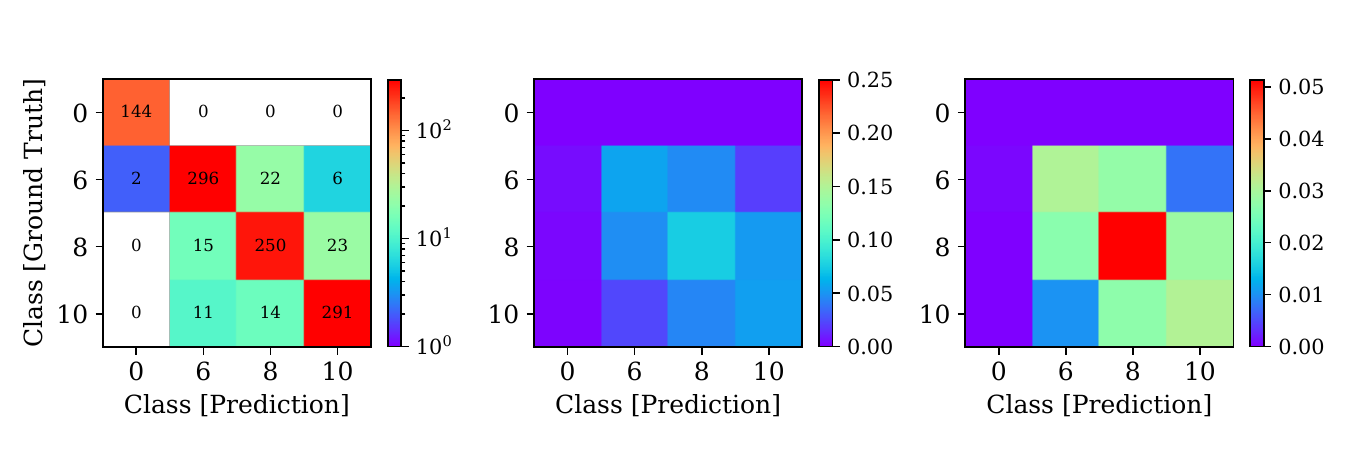}
        \subcaption{Classification of signal power.}
        \label{figure_classification_results2}
    \end{minipage}
	\begin{minipage}[t]{1.0\linewidth}
        \centering
        \includegraphics[trim=10 14 12 34, clip, width=1.0\linewidth]{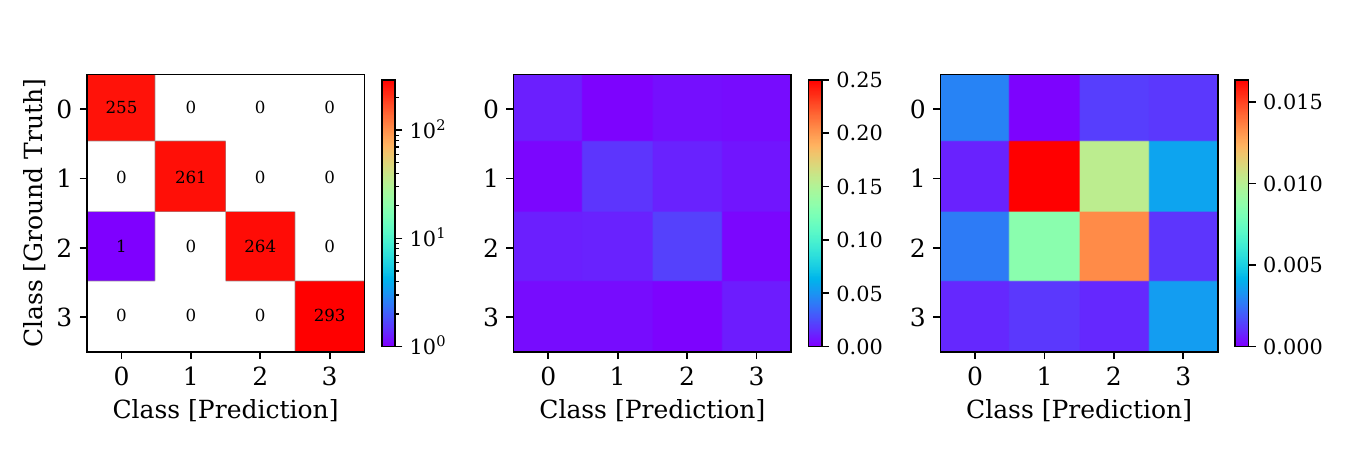}
        \subcaption{Classification of antenna area.}
        \label{figure_classification_results3}
    \end{minipage}
    \caption{Classification results: Confusion matrix (left), aleatoric uncertainty (middle), and epistemic uncertainty (right) for snapshot input length of 34.}
    \label{figure_classification_results}
\end{figure}

\textbf{Interference Type.} The model attains a classification accuracy of 99.88\% in distinguishing interferences, with only a solitary misclassification evident (see the confusion matrix depicted in Figure~\ref{figure_classification_results1}). The aleatoric uncertainty manifests an equitable distribution due to the availability of all necessary information within snapshots of a length of 34. Nonetheless, the model encounters challenges in discerning classes 2, 3, 4, and 6, due to similar snapshots, which results in higher epistemic uncertainty.

\textbf{Signal Power.} The model yields an accuracy of 86.42\% on the signal power classification task, yet it exhibits misclassifications between signal power $6\,\text{dBm}$ and $8\,\text{dBm}$, as well as between $8\,\text{dBm}$ and $10\,\text{dBm}$ (see Figure~\ref{figure_classification_results2}). However, detecting no interference remains reliable. The model's capability to accurately distinguish between various signal powers is hindered by an increase of aleatoric and epistemic uncertainty. An alternative normalization procedure could potentially amplify the distinctions among snapshots. The regression task poses a challenge for the model, resulting in an error of $0.81\,\text{dBm}$. Although the classification accuracy marginally reduces to 96.15\% in an independent task, the accuracy on the signal power drops significantly to 55.96\% (refer to Table~\ref{table_all_results}), due to the intricate generalization toward unfamiliar signal powers.

\textbf{Antenna Area.} Although the model demonstrates near-perfect classification capability in distinguishing antenna areas, achieving 98.26\%, notable uncertainty persists, particularly concerning the differentiation between area 1 and 2 (see Figure~\ref{figure_classification_results3}). This uncertainty primarily stems from the minimal spatial separation of merely $7\,cm$ between the respective areas.

\newcommand\ac{0.325}
\begin{figure}[!t]
    \centering
	\begin{minipage}[t]{\ac\linewidth}
        \centering
    	\includegraphics[trim=10 10 10 10, clip, width=1.0\linewidth]{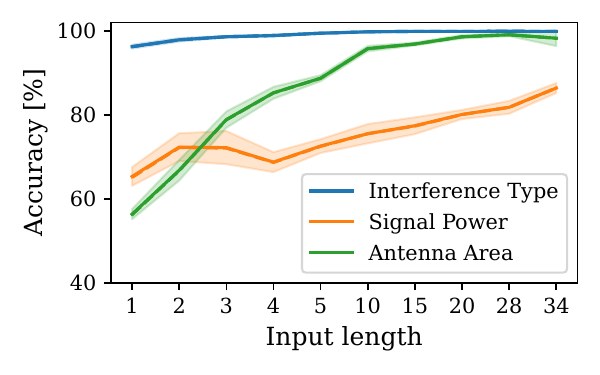}
        \vspace{-0.5cm}
    	\subcaption{Dependent task.}
    	\label{figure_snapshot_length1}
    \end{minipage}
    \hfill
	\begin{minipage}[t]{\ac\linewidth}
        \centering
    	\includegraphics[trim=10 10 10 10, clip, width=1.0\linewidth]{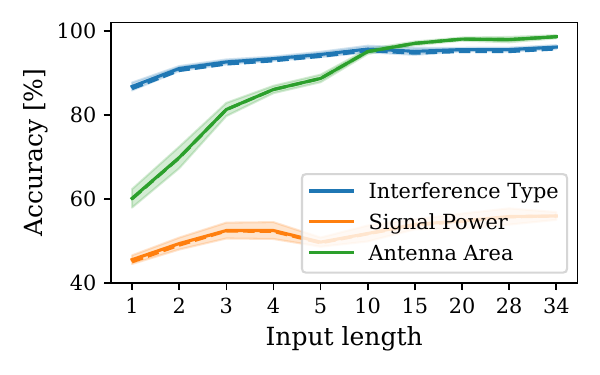}
        \vspace{-0.5cm}
    	\subcaption{Independent task.}
    	\label{figure_snapshot_length2}
    \end{minipage}
    \hfill
	\begin{minipage}[t]{\ac\linewidth}
        \centering
    	\includegraphics[trim=10 10 10 10, clip, width=1.0\linewidth]{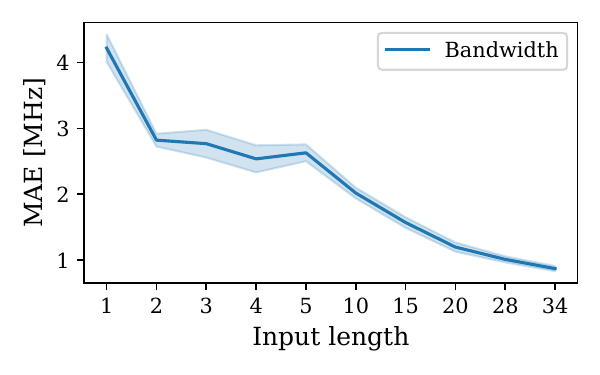}
        \vspace{-0.5cm}
    	\subcaption{Bandwidth.}
    	\label{figure_snapshot_length3}
    \end{minipage}
	\begin{minipage}[t]{\ac\linewidth}
        \centering
    	\includegraphics[trim=10 10 10 10, clip, width=1.0\linewidth]{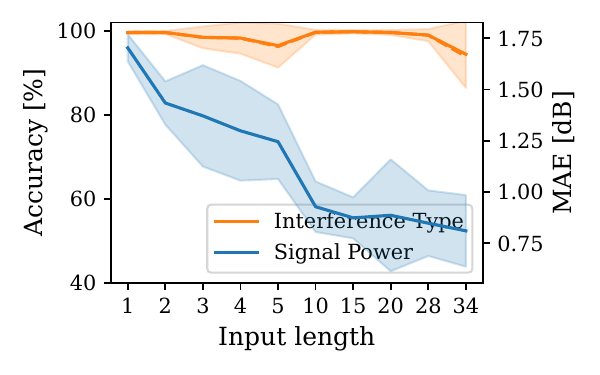}
        \vspace{-0.5cm}
    	\subcaption{Signal power.}
    	\label{figure_snapshot_length4}
    \end{minipage}
    \hfill
	\begin{minipage}[t]{\ac\linewidth}
        \centering
    	\includegraphics[trim=10 10 10 10, clip, width=1.0\linewidth]{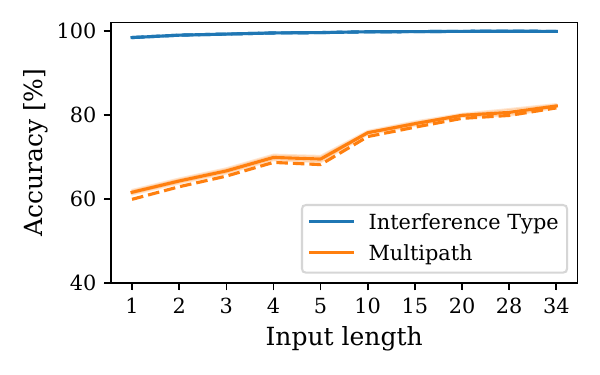}
        \vspace{-0.5cm}
    	\subcaption{Scenario.}
    	\label{figure_snapshot_length5}
    \end{minipage}
    \hfill
	\begin{minipage}[t]{\ac\linewidth}
        \centering
    	\includegraphics[trim=10 10 10 10, clip, width=1.0\linewidth]{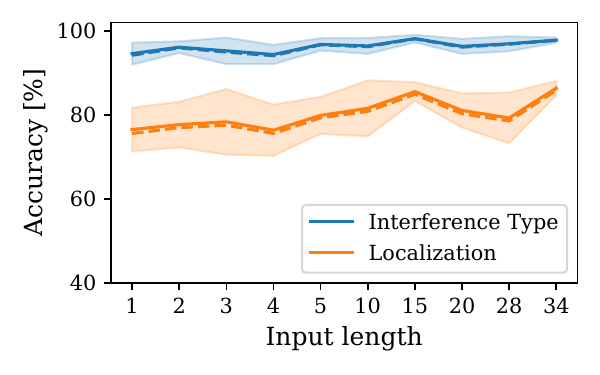}
        \vspace{-0.5cm}
    	\subcaption{Position area.}
    	\label{figure_snapshot_length6}
    \end{minipage}
	\begin{minipage}[t]{\ac\linewidth}
        \centering
    	\includegraphics[trim=10 10 10 10, clip, width=1.0\linewidth]{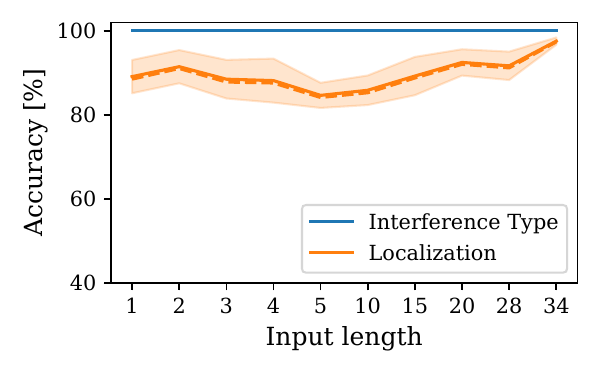}
        \vspace{-0.5cm}
    	\subcaption{Position area$^{\text{1}}$.}
    	\label{figure_snapshot_length7}
    \end{minipage}
    \hfill
	\begin{minipage}[t]{\ac\linewidth}
        \centering
    	\includegraphics[trim=10 10 10 10, clip, width=1.0\linewidth]{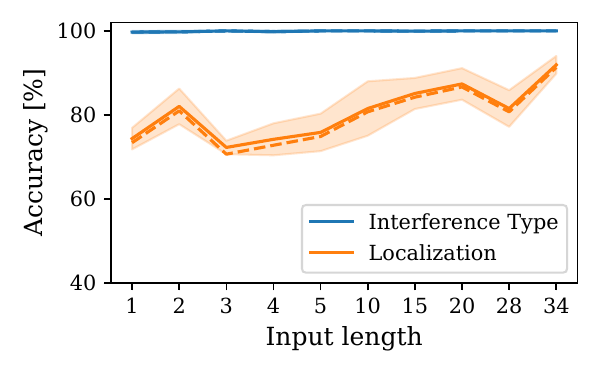}
        \vspace{-0.5cm}
    	\subcaption{Position gallery.}
    	\label{figure_snapshot_length8}
    \end{minipage}
    \hfill
	\begin{minipage}[t]{\ac\linewidth}
        \centering
    	\includegraphics[trim=10 10 10 10, clip, width=1.0\linewidth]{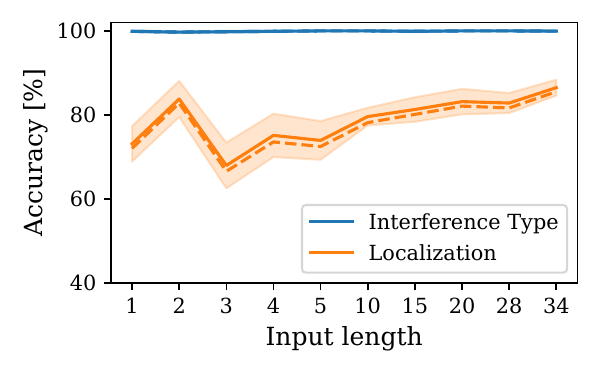}
        \vspace{-0.5cm}
    	\subcaption{Area$^{\text{1}}$ + gallery.}
    	\label{figure_snapshot_length9}
    \end{minipage}
    \caption{Evaluation results for various combinations of classification and regressions tasks (F2-score as dashed line). $^{\text{1}}$\footnotesize{Only overlapping sub-interferences.}}
    \label{figure_snapshot_length}
\end{figure}

\textbf{Bandwidth.} We conduct regression analysis on the BW, yielding an error of $0.87\,\text{MHz}$. This error remains low, owing to the large BW range spanning from $0.2\,\text{MHz}$ to $60.0\,\text{MHz}$ (refer to Table~\ref{table_bandwidth}). Therefore, the model is capable of accurately predicting BW across a wide range (even when deployed in outdoor environments).

\textbf{Interference Localization.} The classification of positions within the open area, achieving an accuracy of 86.38\%, as well as those on the gallery (91.93\%), is notably robust. However, to train a regression task necessitates a substantially larger amount of snapshots from diverse spread positions.

\textbf{Analysis of Snapshot Length.} In Figure~\ref{figure_snapshot_length}, we conduct an analysis of task accuracy relative to snapshot length. Concerning interference classification, accuracy exhibits only marginal decline at lower lengths, i.e., when $\text{N}_{\text{t}} = 1$ (see Figure~\ref{figure_snapshot_length1}). However, for the antenna area task, robust performance is evident only for signal lengths higher than $\text{N}_{\text{t}} = 5$, while falling below 70\% for the signal power estimation. Regarding BW estimation, a complete length of $\text{N}_{\text{t}} = 34$ is imperative (see Figure~\ref{figure_snapshot_length3}), a requirement similarly observed in the signal power regression task (see Figure~\ref{figure_snapshot_length4}). The model demonstrates robustness across all input lengths for the four localization tasks (see Figure~\ref{figure_snapshot_length6} to \ref{figure_snapshot_length9}). Hence, there exists potential to reduce the snapshot length to approximately $\text{N}_{\text{t}} = 10$, while still maintaining high accuracies.

\newcommand\cf{0.325}
\begin{figure}[!t]
    \centering
	\begin{minipage}[t]{\cf\linewidth}
        \centering
        \includegraphics[trim=10 20 6 34, clip, width=1.0\linewidth]{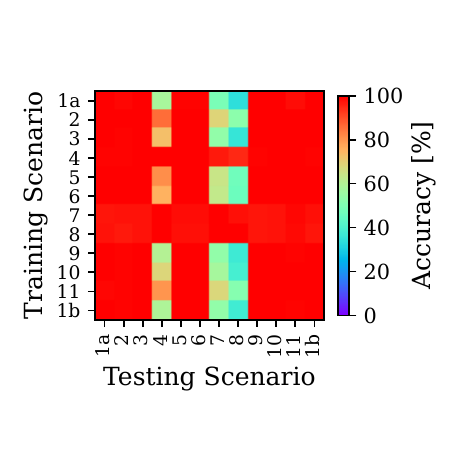}
        \subcaption{Interference type.}
        \label{figure_scenarios_cross1}
    \end{minipage}
    \hfill
	\begin{minipage}[t]{\cf\linewidth}
        \centering
        \includegraphics[trim=10 20 6 34, clip, width=1.0\linewidth]{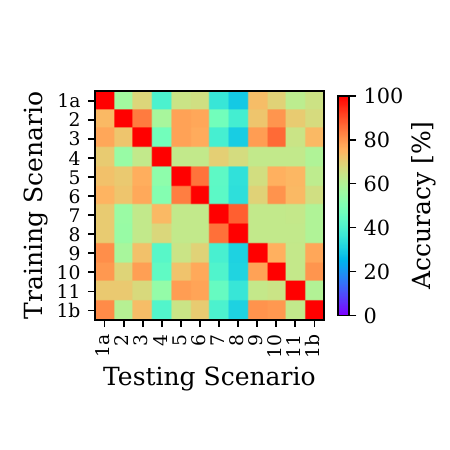}
        \subcaption{Signal power.}
        \label{figure_scenarios_cross2}
    \end{minipage}
    \hfill
	\begin{minipage}[t]{\cf\linewidth}
        \centering
        \includegraphics[trim=10 20 6 34, clip, width=1.0\linewidth]{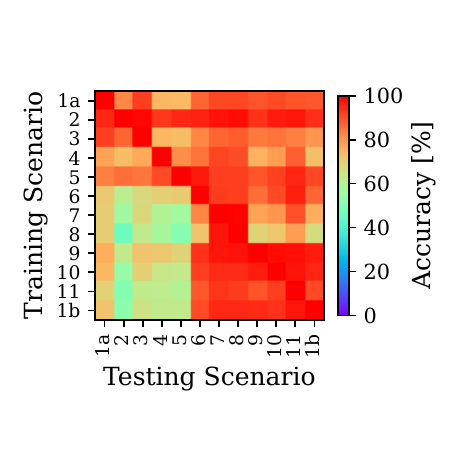}
        \subcaption{Antenna area.}
        \label{figure_scenarios_cross3}
    \end{minipage}
    \caption{Classification results for cross-validating all 11 scenarios. We differentiate 1a and 1b by their antenna and MXG positions in the open scenario.}
    \label{figure_scenarios_cross}
\end{figure}

\begin{figure*}[!t]
    \centering
    \includegraphics[trim=10 10 10 10, clip, width=1.0\linewidth]{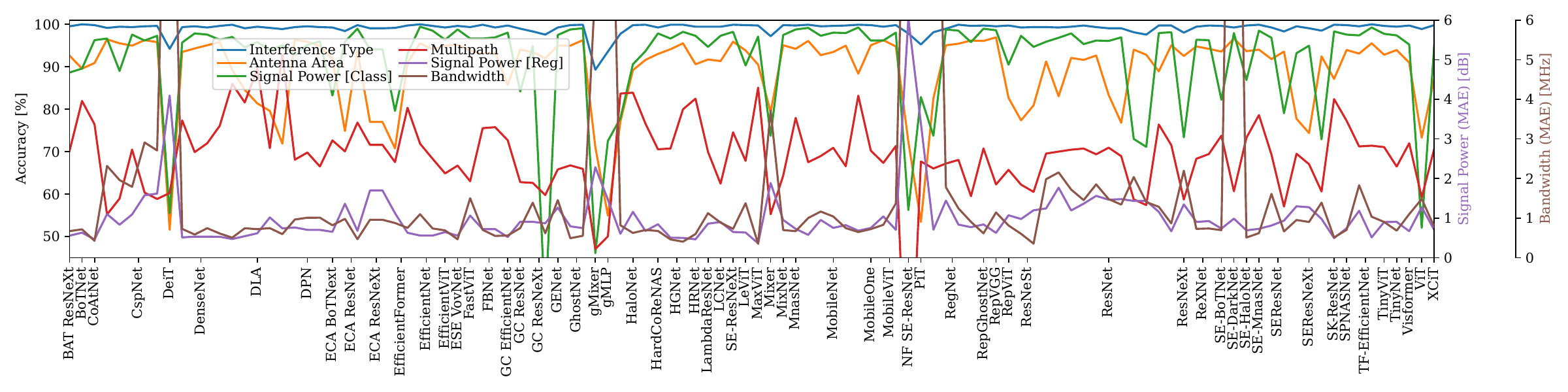}
    \caption{Evaluation of the Hugging Face image models for the classification tasks (interference type, antenna area, signal power, and multipath scenario in \%) and the regression tasks (signal power in dBm and bandwidth in MHz evaluated with the MAE metric). The omitted x-axis tick labels represent variations in the model sizes.}
    \label{figure_results_timm}
\end{figure*}

\textbf{Multipath.} When combining all 11 scenarios, the model achieves a multipath classification accuracy of 82.14\%. This outcome can be explained by the comprehensive cross-validation across all scenarios (refer to Figure~\ref{figure_scenarios_cross}). Notably, a considerable performance decline is observed in scenarios 4, 7, and 8 (falling below 40\%, refer to Table~\ref{table_results_multi_path}), attributed to complete signal absorption within these scenarios (see Figures~\ref{figure_multi_path4}, \ref{figure_multi_path7}, and \ref{figure_multi_path8}). Successful classification of the signal power is feasible solely under conditions of equal train and test scenarios (refer to Figure~\ref{figure_scenarios_cross2}), owing to the comparable influence of multipath effects and various signal powers. Confusion arises in the classification of antenna area due to multipath effects, evident across scenarios 1 to 5 and 6 to 11 (refer to Figure~\ref{figure_scenarios_cross3}). Evaluations conducted in a real-world outdoor environment, free from multipath effects, demonstrated that an ML model trained on indoor data can accurately predict interference and BW classifications in outdoor settings (not shown here as it is not within this topic), see also \cite{heublein_feigl_posnav}.

\begin{table}[t!]
\begin{center}
\setlength{\tabcolsep}{1.0pt}
    \caption{Results (mean accuracy in \%) for cross-validating multipath scenarios.}
    \label{table_results_multi_path}
    \scriptsize \begin{tabular}{ p{1.05cm} | p{0.5cm} | p{0.5cm} | p{0.5cm} | p{0.5cm} | p{0.5cm} | p{0.5cm} | p{0.5cm} | p{0.5cm} | p{0.5cm} | p{0.5cm} | p{0.5cm} | p{0.5cm} }
    \diagbox[innerwidth=4.3em]{\textbf{Train}}{\textbf{Test}} & \multicolumn{1}{c|}{\textbf{1a}} & \multicolumn{1}{c|}{\textbf{2}} & \multicolumn{1}{c|}{\textbf{3}} & \multicolumn{1}{c|}{\textbf{4}} & \multicolumn{1}{c|}{\textbf{5}} & \multicolumn{1}{c|}{\textbf{6}} & \multicolumn{1}{c|}{\textbf{7}} & \multicolumn{1}{c|}{\textbf{8}} & \multicolumn{1}{c|}{\textbf{9}} & \multicolumn{1}{c|}{\textbf{10}} & \multicolumn{1}{c|}{\textbf{11}} & \multicolumn{1}{c}{\textbf{1b}} \\ \hline
    \multicolumn{1}{r|}{\textbf{1a}} & \multicolumn{1}{r|}{100.0} & \multicolumn{1}{r|}{99.19} & \multicolumn{1}{r|}{99.93} & \multicolumn{1}{r|}{58.16} & \multicolumn{1}{r|}{99.59} & \multicolumn{1}{r|}{99.33} & \multicolumn{1}{r|}{48.94} & \multicolumn{1}{r|}{34.14} & \multicolumn{1}{r|}{99.98} & \multicolumn{1}{r|}{99.97} & \multicolumn{1}{r|}{98.06} & \multicolumn{1}{r}{99.99} \\
    \multicolumn{1}{r|}{\textbf{2}} & \multicolumn{1}{r|}{99.96} & \multicolumn{1}{r|}{100.0} & \multicolumn{1}{r|}{100.0} & \multicolumn{1}{r|}{85.69} & \multicolumn{1}{r|}{100.0} & \multicolumn{1}{r|}{99.98} & \multicolumn{1}{r|}{68.73} & \multicolumn{1}{r|}{52.60} & \multicolumn{1}{r|}{100.0} & \multicolumn{1}{r|}{100.0} & \multicolumn{1}{r|}{99.96} & \multicolumn{1}{r}{100.0} \\
    \multicolumn{1}{r|}{\textbf{3}} & \multicolumn{1}{r|}{100.0} & \multicolumn{1}{r|}{99.55} & \multicolumn{1}{r|}{100.0} & \multicolumn{1}{r|}{72.84} & \multicolumn{1}{r|}{99.99} & \multicolumn{1}{r|}{99.94} & \multicolumn{1}{r|}{53.66} & \multicolumn{1}{r|}{35.83} & \multicolumn{1}{r|}{99.99} & \multicolumn{1}{r|}{100.0} & \multicolumn{1}{r|}{99.82} & \multicolumn{1}{r}{100.0} \\
    \multicolumn{1}{r|}{\textbf{4}} & \multicolumn{1}{r|}{99.37} & \multicolumn{1}{r|}{99.59} & \multicolumn{1}{r|}{99.64} & \multicolumn{1}{r|}{100.0} & \multicolumn{1}{r|}{99.80} & \multicolumn{1}{r|}{99.82} & \multicolumn{1}{r|}{96.68} & \multicolumn{1}{r|}{94.67} & \multicolumn{1}{r|}{99.45} & \multicolumn{1}{r|}{99.78} & \multicolumn{1}{r|}{99.95} & \multicolumn{1}{r}{99.57} \\
    \multicolumn{1}{r|}{\textbf{5}} & \multicolumn{1}{r|}{99.97} & \multicolumn{1}{r|}{99.70} & \multicolumn{1}{r|}{100.0} & \multicolumn{1}{r|}{81.20} & \multicolumn{1}{r|}{100.0} & \multicolumn{1}{r|}{100.0} & \multicolumn{1}{r|}{64.18} & \multicolumn{1}{r|}{46.94} & \multicolumn{1}{r|}{100.0} & \multicolumn{1}{r|}{100.0} & \multicolumn{1}{r|}{99.97} & \multicolumn{1}{r}{100.0} \\
    \multicolumn{1}{r|}{\textbf{6}} & \multicolumn{1}{r|}{99.99} & \multicolumn{1}{r|}{99.68} & \multicolumn{1}{r|}{100.0} & \multicolumn{1}{r|}{75.35} & \multicolumn{1}{r|}{100.0} & \multicolumn{1}{r|}{100.0} & \multicolumn{1}{r|}{63.02} & \multicolumn{1}{r|}{46.37} & \multicolumn{1}{r|}{100.0} & \multicolumn{1}{r|}{100.0} & \multicolumn{1}{r|}{99.99} & \multicolumn{1}{r}{100.0} \\
    \multicolumn{1}{r|}{\textbf{7}} & \multicolumn{1}{r|}{96.92} & \multicolumn{1}{r|}{97.40} & \multicolumn{1}{r|}{97.63} & \multicolumn{1}{r|}{99.80} & \multicolumn{1}{r|}{98.40} & \multicolumn{1}{r|}{98.36} & \multicolumn{1}{r|}{100.0} & \multicolumn{1}{r|}{97.75} & \multicolumn{1}{r|}{97.07} & \multicolumn{1}{r|}{97.54} & \multicolumn{1}{r|}{98.90} & \multicolumn{1}{r}{97.78} \\
    \multicolumn{1}{r|}{\textbf{8}} & \multicolumn{1}{r|}{97.31} & \multicolumn{1}{r|}{96.68} & \multicolumn{1}{r|}{97.31} & \multicolumn{1}{r|}{99.83} & \multicolumn{1}{r|}{97.87} & \multicolumn{1}{r|}{97.76} & \multicolumn{1}{r|}{100.0} & \multicolumn{1}{r|}{100.0} & \multicolumn{1}{r|}{97.10} & \multicolumn{1}{r|}{97.57} & \multicolumn{1}{r|}{98.69} & \multicolumn{1}{r}{97.09} \\
    \multicolumn{1}{r|}{\textbf{9}} & \multicolumn{1}{r|}{100.0} & \multicolumn{1}{r|}{99.41} & \multicolumn{1}{r|}{99.99} & \multicolumn{1}{r|}{60.23} & \multicolumn{1}{r|}{99.70} & \multicolumn{1}{r|}{99.96} & \multicolumn{1}{r|}{53.79} & \multicolumn{1}{r|}{37.22} & \multicolumn{1}{r|}{100.0} & \multicolumn{1}{r|}{100.0} & \multicolumn{1}{r|}{99.58} & \multicolumn{1}{r}{100.0} \\
    \multicolumn{1}{r|}{\textbf{10}} & \multicolumn{1}{r|}{100.0} & \multicolumn{1}{r|}{99.56} & \multicolumn{1}{r|}{100.0} & \multicolumn{1}{r|}{68.28} & \multicolumn{1}{r|}{99.97} & \multicolumn{1}{r|}{99.96} & \multicolumn{1}{r|}{57.65} & \multicolumn{1}{r|}{38.89} & \multicolumn{1}{r|}{99.99} & \multicolumn{1}{r|}{100.0} & \multicolumn{1}{r|}{99.84} & \multicolumn{1}{r}{100.0} \\
    \multicolumn{1}{r|}{\textbf{11}} & \multicolumn{1}{r|}{99.19} & \multicolumn{1}{r|}{99.59} & \multicolumn{1}{r|}{99.89} & \multicolumn{1}{r|}{79.82} & \multicolumn{1}{r|}{100.0} & \multicolumn{1}{r|}{100.0} & \multicolumn{1}{r|}{67.84} & \multicolumn{1}{r|}{51.44} & \multicolumn{1}{r|}{99.72} & \multicolumn{1}{r|}{99.98} & \multicolumn{1}{r|}{100.0} & \multicolumn{1}{r}{99.66} \\
    \multicolumn{1}{r|}{\textbf{1b}} & \multicolumn{1}{r|}{100.0} & \multicolumn{1}{r|}{99.47} & \multicolumn{1}{r|}{100.0} & \multicolumn{1}{r|}{59.22} & \multicolumn{1}{r|}{99.84} & \multicolumn{1}{r|}{99.73} & \multicolumn{1}{r|}{53.86} & \multicolumn{1}{r|}{37.77} & \multicolumn{1}{r|}{99.99} & \multicolumn{1}{r|}{100.0} & \multicolumn{1}{r|}{99.37} & \multicolumn{1}{r}{100.0} \\
    \end{tabular}
\end{center}
\end{table}

\begin{figure}[!t]
    \centering
    \includegraphics[trim=10 10 10 10, clip, width=1.0\linewidth]{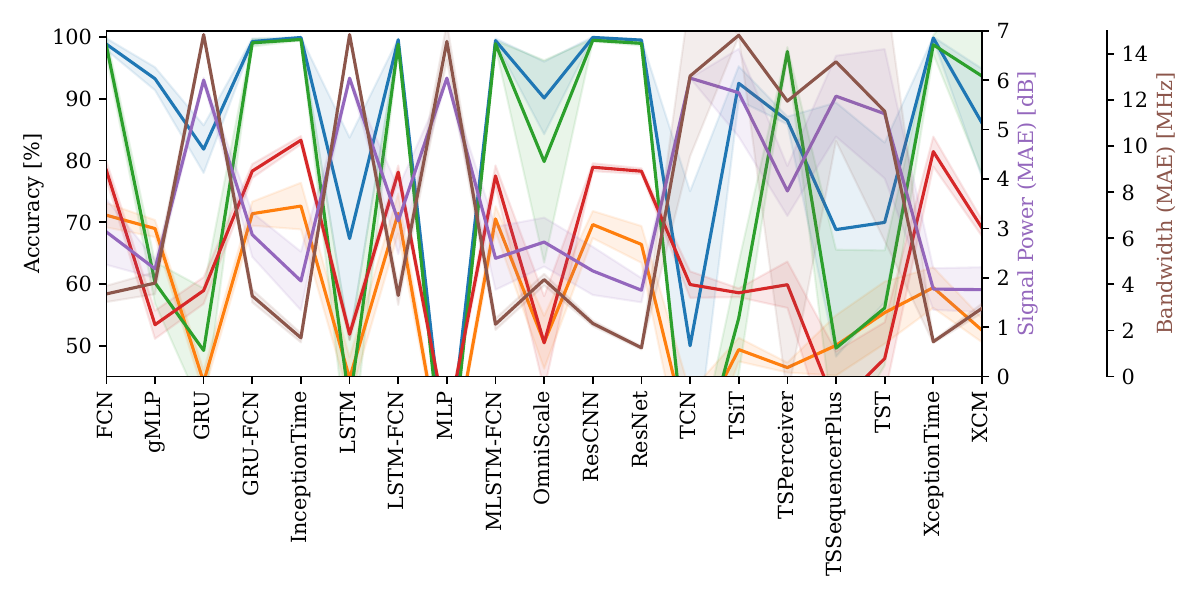}
    \caption{Evaluation of tsai models; comparable with Figure~\ref{figure_results_timm}.}
    \label{figure_results_tsai}
\end{figure}

\textbf{Architecture Benchmark.} Figure~\ref{figure_results_tsai} presents results of 19 tsai models for all characterization tasks. The plot includes shaded regions representing standard deviations, which highlight the uncertainty or variability in the models' predictions. ResNet and ResCNN, along with smaller models such as FCN, GRU-FCN, and InceptionTime, achieve high classification accuracies for both interference and signal power. However, ResNet consistently outperforms all other models across all tasks. Figure~\ref{figure_results_timm} benchmarks 110 Hugging Face models, revealing varying accuracy trends, with some models maintaining consistently high performance while others exhibit significant fluctuations. The best-performing models include ResNet variants, EfficientNet, and ConvNeXt, which demonstrate strong and stable accuracy, making them the most reliable choices for interference characterization and related tasks.

%% file: 06conclusion.tex
\section{Conclusion}
\label{label_conclusion}

We recorded a large snapshot-based GNSS dataset in a controlled indoor environment. This dataset serves as the foundation for evaluating the robustness of ML techniques for interference classification, characterization, and localization tasks derived from a signal generator. We captured diverse scenarios for classification of interference type, signal power, bandwidth, antenna area, localization, and multipath scenario. Through meticulous analysis involving uncertainty computation, snapshot length, and a benchmark of 129 vision models, we demonstrated that the ResNet18 model exhibits resilience under challenging setups, even when subjected to brief input lengths. Owing to the significant influence on the performance of ML models, including the diverse array of potential (hardware) jammer types with distinct interference patterns, antenna properties, as well as environmental dynamics, necessitates the exploration of distinct research pathways, such as data augmentation, transfer learning, continual learning, and federated learning.